\documentclass[10pt,twocolumn,letterpaper]{article}
\usepackage[pagenumbers]{cvpr}
\usepackage{graphicx}
\usepackage{amsmath}
\usepackage{amssymb}
\usepackage{booktabs}

\usepackage{enumitem}
\usepackage{xcolor}

\definecolor{ForestGreen}{rgb}{0.13, 0.55, 0.13}
\usepackage{pifont}
\newcommand{\cmark}{{\color{ForestGreen} \ding{51}}}%
\newcommand{\xmark}{{\color{red} \ding{55}}}%
\usepackage{booktabs}

\usepackage{colortbl}
\usepackage{tabulary}
\usepackage{etoolbox}
\usepackage{multirow}

\usepackage{enumitem}
\setitemize{noitemsep,topsep=0pt,parsep=0pt,partopsep=0pt}
\setenumerate{noitemsep,topsep=0pt,parsep=0pt,partopsep=0pt}
\usepackage[pagebackref,breaklinks,colorlinks]{hyperref}
\usepackage[percentage]{overpic}

\usepackage[capitalize]{cleveref}
\crefname{section}{Sec.}{Secs.}
\Crefname{section}{Section}{Sections}
\Crefname{table}{Table}{Tables}
\crefname{table}{Tab.}{Tabs.}

\def\cvprPaperID{4298} 
\def\confName{CVPR}
\def\confYear{2023}

\begin{document}

\title{
Self-Supervised Learning for Multimodal Non-Rigid 3D Shape Matching 
}

\author{Dongliang Cao $\qquad$ Florian Bernard\\
University of Bonn
}
\maketitle

\begin{abstract}
The matching of 3D shapes has been extensively studied for shapes represented as surface meshes, as well as for shapes represented as point clouds. While point clouds are a common representation of raw real-world 3D data (e.g.~from laser scanners), meshes encode rich and expressive topological information, but their creation typically requires some form of (often manual) curation. In turn, methods that purely rely on point clouds are unable to meet the matching quality of mesh-based methods that utilise the additional topological structure. In this work we close this gap by introducing a self-supervised multimodal learning strategy that combines mesh-based functional map regularisation with a contrastive loss that couples mesh and point cloud data. Our shape matching approach allows to obtain intramodal correspondences for triangle meshes, complete point clouds, and partially observed point clouds, as well as correspondences across these data modalities. We demonstrate that our method achieves state-of-the-art results on several challenging benchmark datasets even in comparison to recent supervised methods, and that our method reaches previously unseen cross-dataset generalisation ability. Our code is available at \scriptsize{\url{https://github.com/dongliangcao/Self-Supervised-Multimodal-Shape-Matching}}.
\end{abstract}

\section{Introduction}
\label{sec:intro}
Matching 3D shapes, i.e.~finding correspondences between their parts, is a fundamental problem in computer vision and computer graphics that has a wide range of applications~\cite{dinh2005texture,loper2015smpl,eisenberger2021neuromorph}. Even though it has been studied for decades~\cite{tam2012registration,van2011survey}, the non-rigid shape matching problem remains highly challenging. One often faces a large variability in terms of shape deformations, or input data with severe noise and topological changes.

\begin{figure}[t!]
\begin{minipage}{0.3\linewidth}
    \centering
    \includegraphics[height=4.0cm]{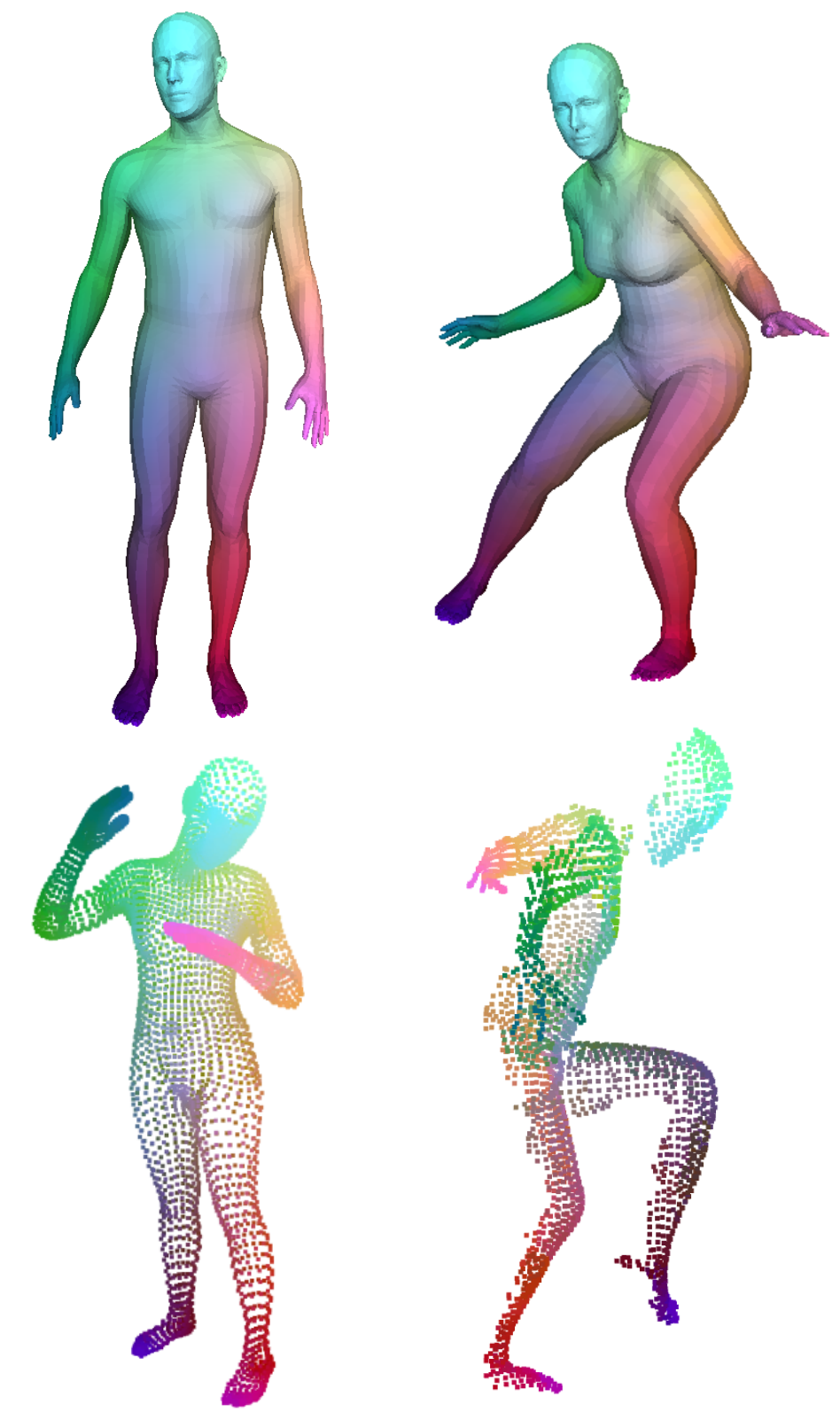}
\end{minipage}
~~~
\begin{minipage}{0.69\linewidth}
    \centering
    \includegraphics[height=4.5cm]{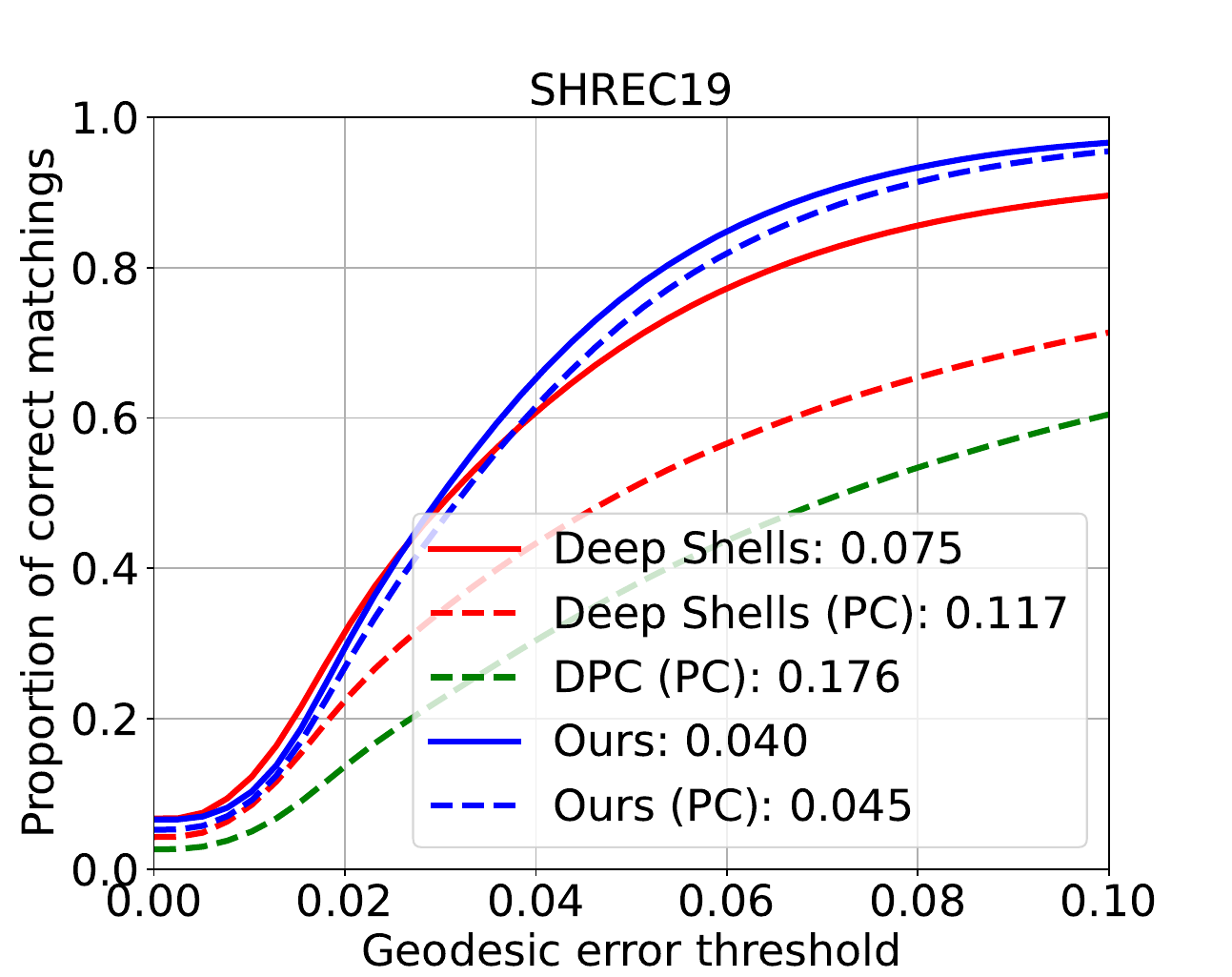}
\end{minipage}
\vspace{-3mm}
\caption{
\textbf{Left:} Our method obtains accurate correspondences  for triangle meshes, point clouds and even partially observed point clouds.
\textbf{Right:} Proportion of correct keypoints (PCK) curves and mean geodesic errors (scores in the legend) on the SHREC'19 dataset~\cite{melzi2019shrec} for meshes (solid lines) and point clouds (dashed lines). Existing point cloud matching methods (DPC~\cite{lang2021dpc}, green line), or mesh-based methods applied to point clouds (Deep Shells~\cite{eisenberger2020deep}, red dashed line) are unable to meet the matching performance of mesh-based methods (solid lines). 
In contrast, our method is multimodal and can process both meshes and point clouds, while enabling accurate shape matching with comparable performance for both modalities (blue lines). }
\label{fig:teaser}
\end{figure}

With the recent success of deep learning, many learning-based approaches were proposed for 3D shape matching~\cite{litany2017deep,masci2015geodesic,groueix20183d,eisenberger2020deep}. 
While recent approaches demonstrate near-perfect matching accuracy without requiring ground truth annotations~\cite{cao2022unsupervised,eisenberger2020deep}, they are limited to 3D shapes represented as triangle meshes and strongly rely on clean data. 

\begin{table*}[tbh!]
\centering
\small
\begin{tabular}{@{}lcccccccc@{}}
\toprule
Methods      & Unsup. & Mesh & Point Cloud & FM-based & Partiality & Robustness & w.o. Refinement &  Required train data${^*}{^*}$ \\ \midrule
FMNet~\cite{litany2017deep} & \xmark &  \cmark & \xmark$^*$ & \cmark & \xmark & \xmark & \cmark & Small \\
GeomFMaps~\cite{donati2020deep} & \xmark &  \cmark & \xmark$^*$ & \cmark & \xmark & \xmark & \cmark & Small \\
DiffFMaps~\cite{marin2020correspondence} & \xmark & \cmark & \cmark & \cmark & \xmark & \cmark & \cmark & Moderate\\
DPFM~\cite{attaiki2021dpfm} & \xmark &  \cmark & \xmark$^*$ & \cmark & \cmark & \xmark & \cmark & Small \\
3D-CODED~\cite{groueix20183d} & \xmark & \cmark & \cmark & \xmark & \cmark & \cmark & \xmark & Large \\
IFMatch~\cite{sundararaman2022implicit} & \xmark & \cmark & \cmark & \xmark & \cmark & \cmark & \xmark & Moderate \\
UnsupFMNet~\cite{halimi2019unsupervised} & \cmark & \cmark & \xmark$^*$ & \cmark & \xmark & \xmark & \cmark & Small \\
SURFMNet~\cite{roufosse2019unsupervised,sharma2020weakly} & \cmark & \cmark & \xmark$^*$ & \cmark & \xmark & \xmark & \cmark & Small \\
Deep Shells~\cite{eisenberger2020deep}  & \cmark & \cmark & \xmark$^*$ & \cmark & \xmark & \xmark & \xmark & Small \\
ConsistFMaps~\cite{cao2022unsupervised} & \cmark & \cmark & \xmark$^*$ & \cmark & \cmark & \xmark & \cmark & Small \\
CorrNet3D~\cite{zeng2021corrnet3d} & \cmark & \cmark & \cmark & \xmark & \xmark & \cmark & \cmark & Large \\
DPC~\cite{lang2021dpc} & \cmark & \cmark & \cmark & \xmark & \xmark & \cmark & \cmark & Moderate \\
Ours & \cmark & \cmark & \cmark & \cmark & \cmark & \cmark & \cmark & Small \\ \bottomrule
\end{tabular}
\vspace{-2mm}
\caption{\textbf{Method comparison.} Our method is the first learning-based approach that combines a unique set of desirable properties. \\
$^*$ Methods are originally designed for meshes, directly applying them to point clouds leads to a large performance drop. \\
${^*}{^*}$ Categorisation according to the amount of training data: \emph{Small {($<$1000)}}, \emph{Moderate {($\approx$5{,}000)}} and \emph{Large {($>$10{,}000)}}.}
\vspace{-3mm}
\label{tab:comparison}
\end{table*}

Since point clouds are a common representation for real-world 3D data, many unsupervised learning approaches were specifically designed for point cloud matching~\cite{groueix2019unsupervised,zeng2021corrnet3d,lang2021dpc}. These methods are often based on learning per-point features, so that  point-wise correspondences are obtained by comparing feature similarities. The learned features were shown to be robust under large shape deformations and severe noise. However, although point clouds commonly represent samples of a surface, respective topological relations are not explicitly available and thus cannot effectively be used during training.
In turn, existing point cloud correspondence methods are unable to meet the matching performance of mesh-based methods, as can be seen  in~\cref{fig:teaser}. Moreover, when applying state-of-the-art unsupervised methods designed for meshes (e.g.~Deep Shells~\cite{eisenberger2020deep}) to point clouds, one can observe a significant drop in matching performance.

In this work, we propose a self-supervised learning framework to address these shortcomings. Our method uses a combination of triangle meshes and point clouds (extracted from the meshes) for training. We first utilise the structural properties of functional maps for triangle meshes as strong unsupervised regularisation. At the same time, we introduce a self-supervised contrastive loss between triangle meshes and corresponding point clouds, enabling the learning of consistent feature representations for both modalities. With that, our method does not require to compute functional maps for point clouds at inference time, but directly predicts correspondences based on feature similarity comparison. 
Overall, our method is the first learning-based approach that combines a unique set of desirable properties, i.e. it can be trained without ground-truth correspondence annotations, is designed for both triangle meshes and point clouds (throughout this paper we refer to this as \emph{multimodal}), is robust against noise, allows for partial shape matching, and requires only a small amount of training data, see \cref{tab:comparison}. In summary, our main contributions are:~
\begin{itemize}[noitemsep,nolistsep]
       \item For the first time we enable \emph{multimodal} non-rigid 3D shape matching under a simple yet efficient \emph{self-supervised learning} framework. 
       \item Our method achieves accurate matchings for triangle meshes based on  \emph{functional map regularisation}, while ensuring matching robustness for less structured point cloud data through  \emph{deep feature similarity}. 
       \item Our method outperforms \emph{state-of-the-art} unsupervised and even supervised methods on several challenging 3D shape matching benchmark datasets and shows previously unseen \emph{cross-dataset generalisation ability}.
       \item
       We extend the SURREAL dataset~\cite{varol2017learning} by SURREAL-PV that exhibits disconnected components in partial views as they occur in 3D scanning scenarios.
\end{itemize}

\section{Related work}
\label{sec:related_work}
Shape matching is a long-standing problem in computer vision and graphics. In the following, we will focus on reviewing those methods that are most relevant to our work. A more comprehensive overview can be found in~\cite{tam2012registration,van2011survey}.

\subsection{Shape matching for triangle meshes}
Triangle meshes are the most common data representation for 3D shapes in computer graphics, thus a large number of matching methods are specifically designed for them~\cite{windheuser2011geometrically,eisenberger2019divergence,sahilliouglu2020recent,bernard2020mina,holzschuh2020simulated,roetzer2022scalable}. Notably, the functional map framework~\cite{ovsjanikov2012functional} is one of the most dominant pipelines in this area and was extended in numerous follow-up works, e.g., in terms of improving the matching accuracy and robustness~\cite{ren2019structured,vestner2017product,melzi2019zoomout}, or by considering non-isometric shape matching~\cite{nogneng2017informative,ren2018continuous,eisenberger2020smooth}, multi-shape matching~\cite{huang2014functional,huang2020consistent,gao2021isometric}, and partial shape matching~\cite{rodola2017partial,litany2017fully}. Meanwhile, with the success of deep learning, many learning-based methods were introduced with the aim to learn improved features compared to handcrafted feature descriptors, such as HKS~\cite{bronstein2010scale}, WKS~\cite{aubry2011wave} or SHOT~\cite{salti2014shot}. FMNet~\cite{litany2017deep} was proposed to learn a vertex-wise non-linear transformation of SHOT descriptors~\cite{salti2014shot}, which is trained in a supervised manner. Later works~\cite{halimi2019unsupervised,roufosse2019unsupervised} enable the unsupervised training of FMNet, and point-based networks~\cite{qi2017pointnet++,wang2019dynamic} were introduced to improve the matching performance~\cite{donati2020deep,sharma2020weakly}. To enable both local and global information propagation, DiffusionNet~\cite{sharp2020diffusionnet} introduced a diffusion layer, which was shown to achieve state-of-the-art performance for 3D shape matching~\cite{attaiki2021dpfm,cao2022unsupervised,donati2022deep,liu2022wtfm}. Even though deep functional map methods were shown to lead to state-of-the-art results for shapes represented as meshes, they are not directly applicable for point cloud matching, since the latter only admit an inaccurate estimation of the Laplace-Beltrami operator (LBO) eigenfunctions~\cite{marin2020correspondence}. To overcome this limitation, our method proposes self-supervised deep feature learning for point clouds without relying on the functional map framework during inference.

\subsection{Shape matching for point clouds}
Point clouds are a commonly used 3D data representation in various real-world applications, e.g., robotics, autonomous driving, AR/VR, etc. Point cloud matching can be roughly classified into two categories: rigid point cloud registration and non-rigid point cloud matching~\cite{tam2012registration}. In this work, we focus on reviewing the learning-based methods for non-rigid point cloud matching. 3D-CODED~\cite{groueix20183d} was proposed to learn a deformation field from a template shape to the given shape to establish correspondences between them. IFMatch~\cite{sundararaman2022implicit} extends vertex-based shape deformation to shape volume deformation to improve the matching robustness. Instead of choosing a template shape beforehand, some works~\cite{groueix2019unsupervised,zeng2021corrnet3d} attempted to learn a pairwise deformation field in an unsupervised manner by shape reconstruction and cycle consistency. However, the introduced deformation network requires a large amount of data to train and is category-specific, which limits the generalisation capability~\cite{lang2021dpc}. In contrast, our method 
requires much less training data and shows previously unseen cross-dataset generalisation ability.

To incorporate the functional map framework into point cloud matching, DiffFMaps~\cite{marin2020correspondence} attempted to learn basis functions with ground truth correspondence supervision to replace the LBO eigenfunctions~\cite{pinkall1993computing}. More recently, DPC~\cite{lang2021dpc} replaced the deformation network by using the shape coordinates themselves to reconstruct shape. Nevertheless, there is still a huge performance gap between unsupervised mesh-based shape matching and point cloud matching~\cite{sundararaman2022implicit}. In this work, we propose a multimodal learning approach to bridge the gap.

\section{Functional maps in a nutshell}
\label{sec:background}\label{subsec:fmaps}
Our approach is based on the functional map framework, which we recap in the following.
Unlike finding point-to-point correspondences, which often leads to NP-hard combinatorial optimisation problems~\cite{lawler1963quadratic}, the functional map framework encodes the correspondence relationship into a small matrix that can be efficiently solved~\cite{ovsjanikov2012functional}. 

\noindent\textbf{Basic pipeline.} Given is a pair of 3D shapes $\mathcal{X}$ and $\mathcal{Y}$ with $n_x$ and $n_y$ vertices, respectively. The functional map framework uses truncated basis functions, i.e.~the first $k$ LBO eigenfunctions~\cite{pinkall1993computing} $\Phi_{x} \in \mathbb{R}^{n_{x} \times k}, \Phi_{y} \in \mathbb{R}^{n_{y} \times k}$, to approximate given features defined on each shape $\mathcal{F}_{x} \in \mathbb{R}^{n_{x} \times c}, \mathcal{F}_{y} \in \mathbb{R}^{n_{y} \times c}$. To this end, the corresponding coefficients $A=\Phi_{x}^{\dagger}\mathcal{F}_{x} \in \mathbb{R}^{k \times c}, B=\Phi_{y}^{\dagger}\mathcal{F}_{y} \in \mathbb{R}^{k \times c}$ are computed for each shape, respectively. The functional map $C_{xy} \in \mathbb{R}^{k \times k}$ can be computed by solving the continuous optimisation problem
        \begin{equation}
            \label{eq:fmap}
            C_{xy}=\mathrm{argmin}_{C}~ E_{\mathrm{data}}\left(C\right)+\lambda E_{\mathrm{reg}}\left(C\right),
        \end{equation}
where minimising $E_{\mathrm{data}}=\left\|CA-B\right\|^{2}$ imposes descriptor preservation, whereas minimising the regularisation term $E_{\mathrm{reg}}$ imposes certain structural properties~\cite{ovsjanikov2012functional}, see \cref{sec:selfsuploss}. From the optimal $C_{xy}$, the point map $\Pi_{yx} \in \{0,1\}^{n_{y} \times n_{x}}$ can be recovered based on the relationship $\Phi_{y}C_{xy} \approx \Pi_{yx}\Phi_{x}$, e.g.~either by nearest neighbour search or by other post-processing techniques~\cite{melzi2019zoomout,pai2021fast,vestner2017product}.

\begin{figure*}
  \centering
  \includegraphics[width=\linewidth]{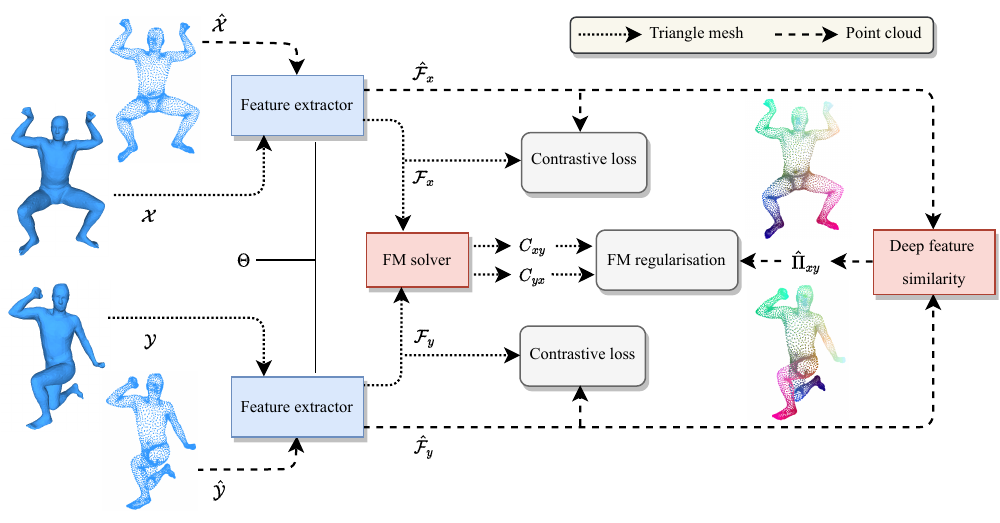}
  \vspace{-9mm}
  \caption{\textbf{Method overview.} During training a Siamese feature extraction network with shared weights $\Theta$ learns to extract mesh features $\mathcal{F}_{x}, \mathcal{F}_{y}$ for input meshes $\mathcal{X}, \mathcal{Y}$, as well as  point cloud features $\mathcal{\hat{F}}_{x}, \mathcal{\hat{F}}_{y}$ for corresponding point clouds $\mathcal{\hat{X}}, \mathcal{\hat{Y}}$. The mesh features $\mathcal{F}_{x}, \mathcal{F}_{y}$ are then used to compute bidirectional functional maps $C_{xy}, C_{yx}$ using the parameter-free FM solver (red). In contrast, the features from point clouds $\mathcal{\hat{F}}_{x}, \mathcal{\hat{F}}_{y}$ are used to construct a soft correspondence matrix $\hat{\Pi}_{xy}$ based on the feature similarity (red). The FM regularisation and contrastive loss together form our overall loss function (gray). {The feature extractor (blue) is the only trainable part in our method.}}
  \label{fig:framework}
\end{figure*}

\section{Non-rigid 3D shape matching}
The whole framework of our approach is depicted in~\cref{fig:framework}. Our approach aims to train a feature extraction network that can be used to extract expressive features for multimodal shape matching. To this end, we pursue a self-supervised training strategy using  multimodal data that comprises meshes and point clouds extracted from these meshes.

To be precise, our multimodal training strategy utilises the shapes $\mathcal{X}$ and $\mathcal{Y}$ represented as triangle meshes, together with corresponding point clouds $\mathcal{\hat{X}}$ and $\mathcal{\hat{Y}}$ that we obtain by discarding the mesh connectivity information and perturbing the vertex coordinates. The same feature extraction network is used to process both  triangle meshes and point clouds, resulting in pointwise features in both cases. Analogous to previous deep functional map methods~\cite{roufosse2019unsupervised,sharma2020weakly,cao2022unsupervised}, a non-trainable FM solver is used to compute bidirectional functional maps $C_{xy}, C_{yx}$ based on the features extracted from the triangle meshes. At the same time, the features extracted from the point clouds are used to construct a soft correspondence matrix $\hat{\Pi}_{xy}$ via feature similarity measurement. To enable the self-supervised training of our feature extractor, we use functional map regularisation. In addition, by using a contrastive loss we enforce that the features from the triangle meshes and the point clouds are similar. 
At inference time, the functional map framework (see~\cref{subsec:fmaps}) is used for finding correspondences for 3D shapes represented as triangle meshes, while the correspondences for point clouds (or between triangle meshes and point clouds) are computed based on deep feature similarity, thereby avoiding the problem of point clouds only admitting an inaccurate estimation of the LBO eigenfunctions~\cite{clarenz2004finite,boscaini2016anisotropic,sharp2020laplacian}. In the following, we explain the individual components of our method in detail.

\subsection{Feature extractor}
The feature extractor aims to extract features of both triangle meshes and point clouds that are robust to shape deformations and to the 
sampling. To this end, we use the DiffusionNet architecture~\cite{sharp2020diffusionnet} throughout our work, similar to other recent methods~\cite{attaiki2021dpfm,cao2022unsupervised}. DiffusionNet is based on an intrinsic surface diffusion process~\cite{sharp2020diffusionnet} and leads to the state-of-the-art performance in the context of shape matching~\cite{attaiki2021dpfm,cao2022unsupervised,donati2022deep,liu2022wtfm}. Moreover, DiffusionNet allows to extract features from both meshes and point clouds.

Following~\cite{cao2022unsupervised}, our feature extractor is used in a Siamese way, i.e. the same network with shared wights $\Theta$ is applied for both source shapes $\mathcal{X}, \hat{\mathcal{X}}$ and target shapes $\mathcal{Y},  \hat{\mathcal{Y}}$.
\subsection{Functional map solver}
\label{subsec:fm_solver}
The goal of the functional map solver (FM solver) is to compute the bidirectional functional maps $C_{xy}, C_{yx}$ based on the extracted features $\mathcal{F}_{x}, \mathcal{F}_{y}$. The basic pipeline is explained in~\cref{subsec:fmaps}. Analogous to previous methods~\cite{attaiki2021dpfm,cao2022unsupervised}, we use a regularised functional map solver~\cite{ren2019structured} to improve the robustness when computing the functional map. To this end, the regularisation term $E_{\mathrm{reg}}$ in~\cref{eq:fmap} can be expressed in the form
    \begin{equation}
        \label{eq:fmap_reg}
        E_{\mathrm{reg}}=\sum_{i j} C_{i j}^{2} M_{i j},
    \end{equation}
where $M_{i j}$ is the resolvent mask that can be viewed as an extension of Laplacian commutativity, see~\cite{ren2019structured} for details.
\subsection{Deep feature similarity}
The goal of the deep feature similarity module is to predict a correspondence matrix $\hat{\Pi}_{xy}$ to explicitly indicate the correspondences between  given input point clouds $\hat{\mathcal{X}}$ and $\hat{\mathcal{Y}}$ with $n_{x}$ and $n_{y}$ points, respectively. Theoretically, $\hat{\Pi}_{xy}$ should be a (partial) permutation matrix, i.e.
    \begin{equation}
        \label{eq:permutation_mat}
        \left\{\Pi \in\{0,1\}^{n_{x} \times n_{y}}: \Pi \mathbf{1}_{n_{y}} \leq \mathbf{1}_{n_{x}}, \mathbf{1}_{n_{x}}^{\top} \Pi \leq \mathbf{1}_{n_{y}}^{\top}\right\},
    \end{equation}
where the element at position $(i,j)$ of $\hat{\Pi}_{xy}$ indicates whether the $i$-th point in $\hat{\mathcal{X}}$ corresponds to the $j$-th point in $\hat{\mathcal{Y}}$. However, the construction of such a binary matrix is non-differentiable. To this end, a soft correspondence matrix is used to approximate the binary constraints~\eqref{eq:permutation_mat} in practice~\cite{eisenberger2021neuromorph,cao2022unsupervised,saleh2022bending,zeng2021corrnet3d}.
The key idea to construct the soft correspondence matrix $\hat{\Pi}_{xy}$ is based on the similarity measurement between features $\mathcal{F}_{x}$ and $\mathcal{F}_{y}$ defined on each shape. The construction process can be expressed in the form
    \begin{equation}
        \label{eq:corr}
        \hat{\Pi}_{xy} = \mathrm{Corr}\left(\langle\mathcal{F}_{x}, \mathcal{F}_{y}\rangle\right),
    \end{equation}
where $\langle\cdot, \cdot\rangle$ is the $(n_x {\times} n_y)$-dimensional matrix of the dot products between pairs of feature vectors  and $\mathrm{Corr(\cdot)}$ is an operator to construct a soft correspondence matrix based on the similarity matrix~\cite{mena2018learning,jang2017categorical}.

In this work, we use Sinkhorn normalisation~\cite{sinkhorn1967concerning,mena2018learning} to construct the soft correspondence matrix. Sinkhorn normalisation iteratively normalises rows and columns of a matrix using the softmax operator. During inference, we quantise $\hat{\Pi}_{xy}$ to a binary matrix.

\subsection{Self-supervised loss}\label{sec:selfsuploss}
To train our feature extractor in a self-supervised manner, we combine  unsupervised functional map regularisation~\cite{roufosse2019unsupervised,sharma2020weakly,cao2022unsupervised} with self-supervised contrastive learning~\cite{xie2020pointcontrast}. Our unsupervised functional map regularisation can be divided into two parts. 

The first part regularises the structural properties of the predicted functional maps $C_{xy}, C_{yx}$. Following~\cite{roufosse2019unsupervised}, the functional map regularisation can be expressed in the form
    \begin{equation}
        \label{eq:fmaps}
        E_{\mathrm{fmap}} = \lambda_{\mathrm{bij}}E_{\mathrm{bij}} + \lambda_{\mathrm{orth}}E_{\mathrm{orth}}.
    \end{equation}
In~\cref{eq:fmaps}, $E_{\mathrm{bij}}$ is the bijectivity constraint to ensure the map from $\mathcal{X}$ through $\mathcal{Y}$ back to $\mathcal{X}$ is the identity map, $E_{\mathrm{orth}}$ represents the orthogonality constraint to prompt a locally area-preserving matching, see~\cite{cao2022unsupervised} for more details. 

The second part regularises the predicted soft correspondence matrix $\hat{\Pi}_{xy}$ based on the relationship $\Phi_{x}C_{yx} \approx \hat{\Pi}_{xy}\Phi_{y}$. Following~\cite{cao2022unsupervised}, our unsupervised loss can be expressed in the form
    \begin{equation}
        \label{eq:align_unsup}
        E_{\mathrm{align}} = \|\Phi_{x}C_{yx}-\hat{\Pi}_{xy}\Phi_{y}\|^{2}_{F}.
    \end{equation}
It is important to note that our correspondence matrix $\hat{\Pi}_{xy}$ is directly predicted based on the deep feature similarity between point clouds. This is in contrast to~\cite{cao2022unsupervised}, where a universe classifier is required to predict shape-to-universe point maps.
In turn, our framework is more efficient and flexible, since we do not rely on the universe classifier and the knowledge of the number of universe vertices.

In addition to  functional map regularisation, we further utilise a self-supervised contrastive loss to encourage similar features for corresponding points from the input mesh and the corresponding point cloud. To this end, we use the PointInfoNCE loss~\cite{xie2020pointcontrast}, which maximises the feature similarity between corresponding points in a given triangle mesh $\mathcal{X}$ and a given point cloud $\hat{\mathcal{X}}$, while at the same time minimising the feature similarity between other points. The loss term can be expressed in the form
    \begin{equation}
        \label{eq:nce}
        E_{\mathrm{nce}} = -\sum_{i=1}^{n_{x}} \log \frac{\exp \left( \langle \mathcal{F}_{x}^{i}, \hat{\mathcal{F}}_{x}^{i} \rangle / \tau \right)}{\sum_{j=1}^{n_{x}} \exp \left( \langle  \mathcal{F}_{x}^{i}, \hat{\mathcal{F}}_{x}^{j} \rangle / \tau \right)},
    \end{equation}
where $\tau$ is a scaling factor. Similarly, $E_{\mathrm{nce}}$ is also applied to both shape $\mathcal{Y}$ and $\hat{\mathcal{Y}}$.
Finally, the overall loss for training is a weighted sum of the individual losses above, i.e. 
    \begin{equation}
        \label{eq:total_loss}
        E_{\mathrm{total}} = E_{\mathrm{fmap}} + \lambda_{\mathrm{align}}E_{\mathrm{align}} + \lambda_{\mathrm{nce}}E_{\mathrm{nce}}.
    \end{equation}

\subsection{Implementation details}
We implement our framework in PyTorch. We use DiffusionNet~\cite{sharp2020diffusionnet} with default settings for our feature extractor. In the context of the FM solver, we set $\lambda=100$ in~\cref{eq:fmap}. For training, we empirically set $\lambda_{\mathrm{bij}} = 1.0, \lambda_{\mathrm{orth}} = 1.0$ in~\cref{eq:fmaps}, $\tau = 0.07$ in~\cref{eq:nce}, and $\lambda_{\mathrm{align}} = 10^{-3}, \lambda_{\mathrm{nce}} = 10$ in~\cref{eq:total_loss}. See the supplementary document for more details.

\begin{table*}[hbt!]
\small
    \centering
        \begin{tabular}{@{}lcccccccccc@{}}
        \toprule
        \multicolumn{1}{c}{\multirow{2}{*}{\textbf{Geo. error ($\times$100)}}}      & \multicolumn{3}{c}{\textbf{FAUST}}   & \multicolumn{3}{c}{\textbf{SCAPE}}  & \multicolumn{3}{c}{\textbf{SHREC'19}}  & \multicolumn{1}{c}{\multirow{2}{*}{\textbf{FM-based}}} \\
        & \multicolumn{1}{c}{Mesh} & \multicolumn{1}{c}{PC} & \multicolumn{1}{c}{Noisy PC} & \multicolumn{1}{c}{Mesh} & \multicolumn{1}{c}{PC} & \multicolumn{1}{c}{Noisy PC} & \multicolumn{1}{c}{Mesh} & \multicolumn{1}{c}{PC} & \multicolumn{1}{c}{Noisy PC} & 
        \\ \midrule
        \multicolumn{11}{c}{Axiomatic Methods} \\
        \multicolumn{1}{l}{BCICP \cite{ren2018continuous}}  & \multicolumn{1}{c}{6.4}  & \multicolumn{1}{c}{-} & \multicolumn{1}{c}{-}  & \multicolumn{1}{c}{11.0} & \multicolumn{1}{c}{-} & \multicolumn{1}{c}{-} & \multicolumn{1}{c}{8.0}    & \multicolumn{1}{c}{-} & \multicolumn{1}{c}{-}  & \multicolumn{1}{c}{\cmark}  \\
        \multicolumn{1}{l}{ZOOMOUT \cite{melzi2019zoomout}} & \multicolumn{1}{c}{6.1}  & \multicolumn{1}{c}{-} & \multicolumn{1}{c}{-}  & \multicolumn{1}{c}{7.5} & \multicolumn{1}{c}{-} & \multicolumn{1}{c}{-} & \multicolumn{1}{c}{7.8}    & \multicolumn{1}{c}{-} & \multicolumn{1}{c}{-}  & \multicolumn{1}{c}{\cmark} \\
        \multicolumn{1}{l}{Smooth Shells \cite{eisenberger2020smooth}} & \multicolumn{1}{c}{2.5}  & \multicolumn{1}{c}{-} & \multicolumn{1}{c}{-}  & \multicolumn{1}{c}{4.7} & \multicolumn{1}{c}{-} & \multicolumn{1}{c}{-} & \multicolumn{1}{c}{7.6}    & \multicolumn{1}{c}{-} & \multicolumn{1}{c}{-}  & \multicolumn{1}{c}{\cmark} 
        \\ \midrule
        \multicolumn{11}{c}{Supervised Methods} \\ 
        \multicolumn{1}{l}{FMNet \cite{litany2017deep}} & \multicolumn{1}{c}{3.1}  & \multicolumn{1}{c}{8.5} & \multicolumn{1}{c}{14.0}  & \multicolumn{1}{c}{9.1} & \multicolumn{1}{c}{15.0} & \multicolumn{1}{c}{21.3} & \multicolumn{1}{c}{10.4}    & \multicolumn{1}{c}{14.3} & \multicolumn{1}{c}{19.1}  & \multicolumn{1}{c}{\cmark} \\
        \multicolumn{1}{l}{3D-CODED \cite{groueix20183d}} & \multicolumn{1}{c}{2.5}  & \multicolumn{1}{c}{2.5} & \multicolumn{1}{c}{2.8}  & \multicolumn{1}{c}{9.8} & \multicolumn{1}{c}{9.8} & \multicolumn{1}{c}{10.0} & \multicolumn{1}{c}{7.7}    & \multicolumn{1}{c}{7.7} & \multicolumn{1}{c}{7.9}  & \multicolumn{1}{c}{\xmark} \\
        \multicolumn{1}{l}{IFMatch~\cite{sundararaman2022implicit}} & \multicolumn{1}{c}{2.6} & \multicolumn{1}{c}{2.6} & \multicolumn{1}{c}{2.7} & \multicolumn{1}{c}{11.0} & \multicolumn{1}{c}{11.0} & \multicolumn{1}{c}{11.2} & \multicolumn{1}{c}{6.5} & \multicolumn{1}{c}{6.5} & \multicolumn{1}{c}{6.6} & \multicolumn{1}{c}{\xmark} \\
        \multicolumn{1}{l}{DiffFMaps \cite{marin2020correspondence}} & \multicolumn{1}{c}{10.5}  & \multicolumn{1}{c}{10.5} & \multicolumn{1}{c}{11.7}  & \multicolumn{1}{c}{23.1} & \multicolumn{1}{c}{23.1} & \multicolumn{1}{c}{22.7} & \multicolumn{1}{c}{18.2}    & \multicolumn{1}{c}{18.2} & \multicolumn{1}{c}{19.4}  & \multicolumn{1}{c}{\cmark} \\
        \multicolumn{1}{l}{GeomFMaps~\cite{donati2020deep}}& \multicolumn{1}{c}{2.6}  & \multicolumn{1}{c}{6.1} & \multicolumn{1}{c}{10.2}  & \multicolumn{1}{c}{3.0} & \multicolumn{1}{c}{7.7} & \multicolumn{1}{c}{13.3} & \multicolumn{1}{c}{4.1}    & \multicolumn{1}{c}{10.6} & \multicolumn{1}{c}{14.6}  & \multicolumn{1}{c}{\cmark} \\
        
        \midrule
        \multicolumn{11}{c}{Unsupervised Methods} \\ 
        \multicolumn{1}{l}{SURFMNet \cite{roufosse2019unsupervised,sharma2020weakly} } & \multicolumn{1}{c}{2.4}  & \multicolumn{1}{c}{6.0} & \multicolumn{1}{c}{13.5}  & \multicolumn{1}{c}{6.0} & \multicolumn{1}{c}{11.3} & \multicolumn{1}{c}{20.1} & \multicolumn{1}{c}{4.8}    & \multicolumn{1}{c}{13.9} & \multicolumn{1}{c}{19.1}  & \multicolumn{1}{c}{\cmark} \\
        \multicolumn{1}{l}{UnsupFMNet \cite{halimi2019unsupervised}}  & \multicolumn{1}{c}{4.8}  & \multicolumn{1}{c}{9.6} & \multicolumn{1}{c}{17.8}  & \multicolumn{1}{c}{9.6} & \multicolumn{1}{c}{11.3} & \multicolumn{1}{c}{15.5} & \multicolumn{1}{c}{11.1}    & \multicolumn{1}{c}{17.3} & \multicolumn{1}{c}{23.8}  & \multicolumn{1}{c}{\cmark} \\
        \multicolumn{1}{l}{Deep Shells \cite{eisenberger2020deep}} & \multicolumn{1}{c}{\textbf{1.7}}  & \multicolumn{1}{c}{6.0} & \multicolumn{1}{c}{11.2}  & \multicolumn{1}{c}{5.3} & \multicolumn{1}{c}{7.8} & \multicolumn{1}{c}{11.1} & \multicolumn{1}{c}{7.5}    & \multicolumn{1}{c}{11.7} & \multicolumn{1}{c}{14.4}  & \multicolumn{1}{c}{\cmark} \\
        \multicolumn{1}{l}{ConsistFMaps \cite{cao2022unsupervised}} & \multicolumn{1}{c}{2.4}  & \multicolumn{1}{c}{11.2} & \multicolumn{1}{c}{16.9}  & \multicolumn{1}{c}{5.1} & \multicolumn{1}{c}{12.3} & \multicolumn{1}{c}{16.4} & \multicolumn{1}{c}{4.2}    & \multicolumn{1}{c}{13.7} & \multicolumn{1}{c}{17.2}  & \multicolumn{1}{c}{\cmark} \\
        \multicolumn{1}{l}{CorrNet3D \cite{zeng2021corrnet3d}}  & \multicolumn{1}{c}{26.5}  & \multicolumn{1}{c}{26.5} & \multicolumn{1}{c}{27.0}  & \multicolumn{1}{c}{37.3} & \multicolumn{1}{c}{37.3} & \multicolumn{1}{c}{36.8} & \multicolumn{1}{c}{33.7}    & \multicolumn{1}{c}{33.7} & \multicolumn{1}{c}{34.0}  & \multicolumn{1}{c}{\xmark} \\
        \multicolumn{1}{l}{DPC \cite{lang2021dpc}}  & \multicolumn{1}{c}{11.6}  & \multicolumn{1}{c}{11.6} & \multicolumn{1}{c}{14.6}  & \multicolumn{1}{c}{16.0} & \multicolumn{1}{c}{16.0} & \multicolumn{1}{c}{18.6} & \multicolumn{1}{c}{17.6}    & \multicolumn{1}{c}{17.6} & \multicolumn{1}{c}{19.4}  & \multicolumn{1}{c}{\xmark} \\
        \multicolumn{1}{l}{Ours}  & \multicolumn{1}{c}{2.0}  & \multicolumn{1}{c}{\textbf{2.4}} & \multicolumn{1}{c}{\textbf{4.4}}  & \multicolumn{1}{c}{\textbf{3.1}} & \multicolumn{1}{c}{\textbf{4.1}} & \multicolumn{1}{c}{\textbf{6.6}} & \multicolumn{1}{c}{\textbf{4.0}}    & \multicolumn{1}{c}{\textbf{4.5}} & \multicolumn{1}{c}{\textbf{5.8}}  & \multicolumn{1}{c}{\cmark} \\\hline
        \end{tabular} 
    \vspace{-2mm}
    \caption{Quantitative results on the FAUST, SCAPE and SHREC'19 datasets in terms of mean geodesic errors. We evaluate all methods on individual dataset for shapes represented as triangle meshes and point clouds. The \textbf{best} results from the unsupervised methods in each column are highlighted. The last column indicates whether the method is based on the functional map framework. Our method outperforms previous unsupervised methods and {bridges the matching performance gap between meshes and point clouds.}}
    \label{tab:complete_shape}
\end{table*}

\section{Experimental results}
\label{sec:experiment}
In this section we demonstrate the advantages of our method for multimodal non-rigid 3D shape matching under different challenging scenarios. 

\subsection{Complete shape matching}
\label{subsec:complete_shape}
\noindent \textbf{Datasets.} We evaluate our method on several standard benchmark datasets, namely FAUST~\cite{bogo2014faust}, SCAPE~\cite{anguelov2005scape} and SHREC'19~\cite{melzi2019shrec} dataset. Instead of the original datasets, we choose the more challenging remeshed versions from~\cite{ren2018continuous,donati2020deep}. The FAUST dataset consists of 100 shapes (10 people in 10 poses), where the evaluation is performed on the last 20 shapes. The SCAPE dataset contains 71 different poses of the same person, where the last 20 shapes are used for evaluation. The SHREC'19 dataset is a more challenging benchmark dataset due to significant variations in mesh connectivity. It comprises 44 shapes and a total of 430 evaluation pairs.

\begin{figure*}[hbt!]
    \centering
    \includegraphics[width=\linewidth]{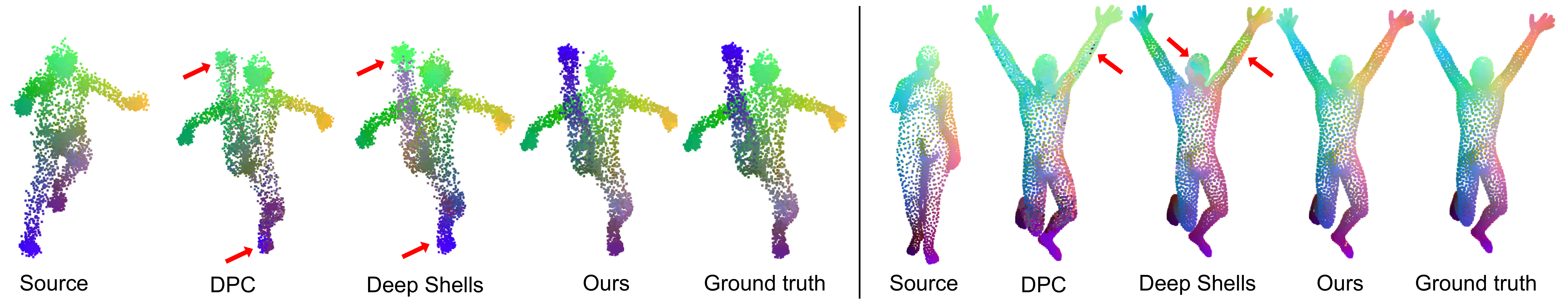}
    \vspace{-7mm}
    \caption{Qualitative results of different methods applied to point clouds from the SHREC'19 dataset. Errors are indicated by red arrows.}
    \label{fig:complete_shape}
\end{figure*}

\noindent \textbf{Results.} The mean geodesic error is used for method evaluation. We compare our method with state-of-the-art axiomatic, supervised and unsupervised methods. To further evaluate the method robustness, we randomly add Gaussian noise to input point clouds. Note that for a fair comparison, we use DiffusionNet~\cite{sharp2020diffusionnet} as the feature extractor for all deep functional map methods, as it can significantly improve shape matching accuracy~\cite{sharp2020diffusionnet}. The results are summarised in~\cref{tab:complete_shape}. We note that directly using DiffusionNet as  feature extractor in previous deep functional map methods does not lead to accurate point cloud matching, see the last column in~\cref{tab:complete_shape}. In contrast, the matching performance of triangle meshes and point clouds remains almost the same for our method. As a result, our method outperforms the previous state-of-the-art in most settings, even in comparison to supervised methods, which is particularly prominent in the case of point cloud matching. \cref{fig:complete_shape} shows a visual comparison of different methods in terms of point cloud matching.

\subsection{Cross-dataset generalisation}
\label{subsec:cross-dataset}

\noindent \textbf{Datasets.} We evaluate the cross-dataset generalisation ability by training on the synthetic SURREAL~\cite{varol2017learning} dataset and evaluating on FAUST, SCAPE and SHREC'19 datasets. Following~\cite{donati2020deep}, we use all (230k) shapes of the SURREAL dataset for point cloud matching methods, while only the first 5k shapes for deep functional map methods, since functional map regularisation is a strong regularisation requiring only a small amount of data to train~\cite{donati2020deep}.

\begin{table}
    \scriptsize
    \centering
        \begin{tabular}{@{}lcccc@{}}
        \toprule
        \multicolumn{1}{c}{\textbf{Geo. ($\times$100)}}      & \multicolumn{1}{c}{\textbf{F (PC)}}   & \multicolumn{1}{c}{\textbf{S (PC)}}  & \multicolumn{1}{c}{\textbf{S19 (PC)}}  & \multicolumn{1}{c}{\textbf{$|$Data$|$}}
        \\ \midrule
        \multicolumn{4}{c}{Supervised Methods} \\ 
        \multicolumn{1}{l}{FMNet~\cite{litany2017deep}}       & \multicolumn{1}{c}{3.8 (12.2)}   & \multicolumn{1}{c}{10.2 (15.3)}     & \multicolumn{1}{c}{13.8 (22.7)} & \multicolumn{1}{c}{5k} \\
        \multicolumn{1}{l}{DiffFMaps~\cite{marin2020correspondence}} & \multicolumn{1}{c}{26.5 (26.5)} & \multicolumn{1}{c}{34.8 (34.8)} & \multicolumn{1}{c}{42.2 (42.2)} & \multicolumn{1}{c}{230k}  \\
        \multicolumn{1}{l}{GeomFMaps~\cite{donati2020deep}}& \multicolumn{1}{c}{2.7 (10.4)} & \multicolumn{1}{c}{3.3 (8.7)}  & \multicolumn{1}{c}{4.7 (14.1)} & \multicolumn{1}{c}{5k}  \\
        
        \midrule
        \multicolumn{4}{c}{Unsupervised Methods}                                                                                           \\ 
        \multicolumn{1}{l}{SURFMNet~\cite{roufosse2019unsupervised,sharma2020weakly}}  & \multicolumn{1}{c}{2.3 (16.0)} & \multicolumn{1}{c}{3.3 (14.7)}  & \multicolumn{1}{c}{8.3 (27.8)} & \multicolumn{1}{c}{5k} \\
        \multicolumn{1}{l}{Deep Shells~\cite{eisenberger2020deep}}  & \multicolumn{1}{c}{8.1 (12.5)}  & \multicolumn{1}{c}{12.2 (14.1)} & \multicolumn{1}{c}{12.1 (15.9)} & \multicolumn{1}{c}{5k} \\
        \multicolumn{1}{l}{ConsistFMaps~\cite{cao2022unsupervised}}  & \multicolumn{1}{c}{3.2 (19.3)}  & \multicolumn{1}{c}{6.7 (17.3)} & \multicolumn{1}{c}{13.7 (24.2)} & \multicolumn{1}{c}{5k} \\
        \multicolumn{1}{l}{CorrNet3D~\cite{zeng2021corrnet3d}}  & \multicolumn{1}{c}{18.1 (18.1)}  & \multicolumn{1}{c}{18.3 (18.3)} & \multicolumn{1}{c}{18.8 (18.8)} & \multicolumn{1}{c}{230k} \\
        \multicolumn{1}{l}{DPC~\cite{lang2021dpc}}  & \multicolumn{1}{c}{13.4 (13.4)}  & \multicolumn{1}{c}{15.8 (15.8)} & \multicolumn{1}{c}{17.4 (17.4)} & \multicolumn{1}{c}{230k} \\
        \multicolumn{1}{l}{Ours}  & \multicolumn{1}{c}{\textbf{2.0} (\textbf{3.5})}  & \multicolumn{1}{c}{\textbf{3.2} (\textbf{3.8})} & \multicolumn{1}{c}{\textbf{4.4} (\textbf{6.6})} & \multicolumn{1}{c}{5k} \\ \hline
        \end{tabular} 
    \vspace{-2mm}
    \caption{Cross-dataset generalisation evaluated on the \textbf{F}AUST, \textbf{S}CAPE and \textbf{S}HREC'\textbf{19} datasets and trained on the SURREAL dataset. The \textbf{best} results in each column are highlighted. The last column indicates the amount of data used for training. Our method outperforms previous supervised and unsupervised methods.}
    \label{tab:generalisation}
\end{table}

\begin{figure}
    \centering
    \includegraphics[width=\linewidth]{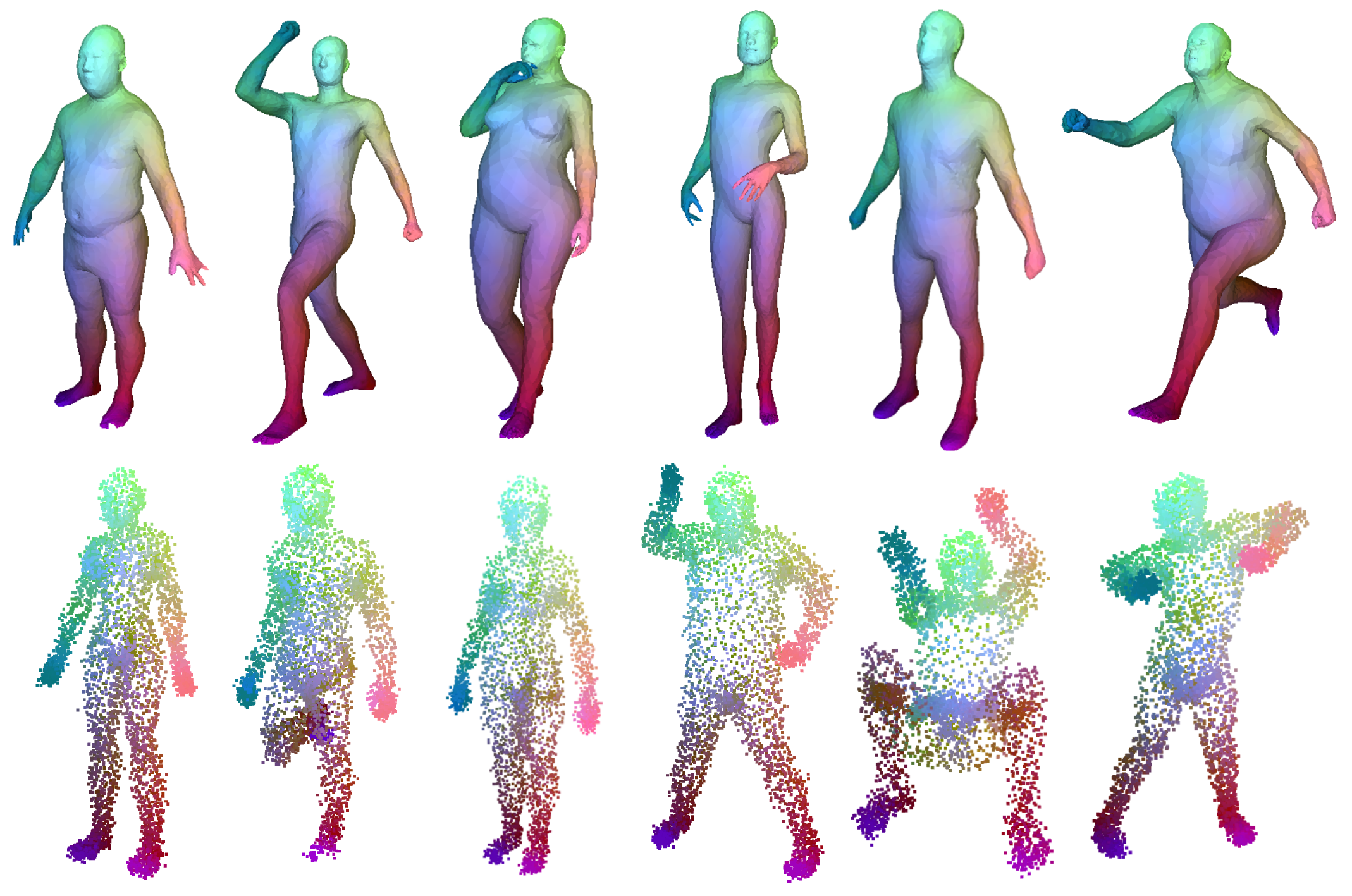}
    \caption{Qualitative results of our method on the SHREC'19 dataset (trained on SURREAL dataset) for both mesh and noisy point cloud matching. The top-left shape is the source shape. {Our method demonstrates previously unseen generalisation ability.}}
    \label{fig:surreal_shrec19}
\end{figure}

\noindent \textbf{Results.} As shown in~\cref{tab:generalisation}, our method achieves a better cross-dataset generalisation ability and outperforms both state-of-the-art supervised and unsupervised methods. We note that deformation-based methods (e.g.~CorrNet3D~\cite{zeng2021corrnet3d}) require a large amount of data to train, since it achieves better results when it is trained on large SURREAL dataset compared to trained on the same small dataset (see comparison between~\cref{tab:complete_shape} and~\cref{tab:generalisation}). In contrast, our method requires only a small amount of training data and achieves similar performance compared to intra-dataset training.~\cref{fig:surreal_shrec19} shows some qualitative results of our method. 

\subsection{Partial shape matching}
\label{subsec:partial_shape}
Our framework can be easily adapted for unsupervised \emph{partial} shape matching, see the supplementary document for details.

\noindent \textbf{Datasets.} We evaluate our method in the context of partial shape matching on the challenging SHREC’16~\cite{cosmo2016shrec} dataset. This dataset consists of 200 training shapes, categorised into 8 classes (humans and animals). Each class has a complete shape to be matched by the other partial shapes. The dataset is
divided into two subsets, namely CUTS (missing a large part) with 120 pairs, and HOLES (missing many small parts) with 80 pairs. Following~\cite{attaiki2021dpfm}, we train our method for each subset individually and evaluate it on the corresponding unseen test set (200 shapes for each subset).

\begin{table}[hbt!]
    \centering
    \footnotesize
        \begin{tabular}{@{}lcc@{}}
        \toprule
        \multicolumn{1}{c}{\textbf{Geo. error ($\times$100)}}      & \multicolumn{1}{c}{\textbf{CUTS (PC)}}   & \multicolumn{1}{c}{\textbf{HOLES (PC)}} 
        \\ \midrule
        \multicolumn{3}{c}{Axiomatic Methods} \\ 
        \multicolumn{1}{l}{PFM~\cite{rodola2017partial}}       & \multicolumn{1}{c}{9.7 (-)}   & \multicolumn{1}{c}{23.2 (-)} \\
        \multicolumn{1}{l}{FSP~\cite{litany2017fully}}       & \multicolumn{1}{c}{16.1 (-)}   & \multicolumn{1}{c}{33.7 (-)} 
        
        \\ \midrule
        \multicolumn{3}{c}{Supervised Methods} \\ 
        \multicolumn{1}{l}{GeomFMaps~\cite{donati2020deep}}& \multicolumn{1}{c}{8.0 (18.5)} & \multicolumn{1}{c}{12.9 (18.9)}   \\
        \multicolumn{1}{l}{DPFM sup~\cite{attaiki2021dpfm}}       & \multicolumn{1}{c}{3.2 (10.4)}   & \multicolumn{1}{c}{11.8 (17.0)}  \\
        
        \midrule
        \multicolumn{3}{c}{Unsupervised Methods}    \\ 
        \multicolumn{1}{l}{ConsistFMaps~\cite{cao2022unsupervised}}  & \multicolumn{1}{c}{8.4 (26.6)}  & \multicolumn{1}{c}{17.9 (27.0)}  \\
        \multicolumn{1}{l}{DPFM unsup~\cite{attaiki2021dpfm}}  & \multicolumn{1}{c}{9.0 (20.9)}  & \multicolumn{1}{c}{20.5 (22.8)}  \\
        \multicolumn{1}{l}{Ours}  & \multicolumn{1}{c}{\textbf{7.6} (\textbf{12.2})}  & \multicolumn{1}{c}{\textbf{15.9} (\textbf{16.7})} \\ \hline
        \end{tabular} 
    \vspace{-2mm}
    \caption{Quantitative results on the CUTS and HOLES subsets of the SHREC’16 dataset. The \textbf{best} results from the unsupervised methods in each column are highlighted. Our method outperforms previous axiomatic and unsupervised methods.}
    \label{tab:partial_comparison}
\end{table}

\begin{figure*}[hbt!]
    \centering
    \includegraphics[width=\linewidth]{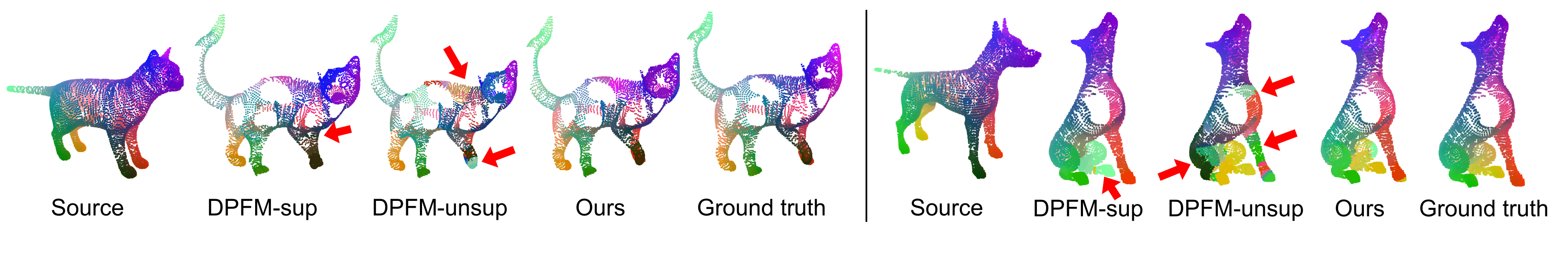}
     \vspace{-11mm}
    \caption{Qualitative results of different methods applied to point clouds from the HOLES subset. Errors are indicated by red arrows.}
    \label{fig:partial_matching}
\end{figure*}

\noindent \textbf{Results.} \cref{tab:partial_comparison} summarises the results. Our method outperforms previous axiomatic and unsupervised methods. Compared to the \emph{supervised} DPFM~\cite{attaiki2021dpfm}, our \emph{self-supervised} method achieves comparable results for point cloud matching.~\cref{fig:partial_matching} shows qualitative results of different methods for point cloud matching on the challenging HOLES subset.

\subsection{Partial view matching}
\label{subsec:partial_view}
In this proof-of-concept experiment we consider the matching of a partially observed point cloud to a complete shape, which is a common scenario for data acquired from 3D scanning devices. To this end, we evaluate our method in terms of partial view matching, in which a partially observed point cloud is  matched to a complete template shape.

\noindent \textbf{Datasets.} We create SURREAL-PV, a new partial view matching dataset based on the SURREAL~\cite{varol2017learning} dataset. Given a complete shape represented as triangle mesh, we use raycasting to obtain a partial shape (both triangle mesh and point cloud) from a randomly sampled viewpoint.
In total, we create 5k shape pairs and divide them into 80\% training set and 20\% test set. Compared to~\cref{subsec:partial_shape}, the challenge of partial view matching is that there exists many disconnected components in the partial shapes and the sampling for partial point clouds is different compared to the complete shapes. 

\begin{table}[hbt!]
    \centering
    \footnotesize
    \begin{tabular}{@{}lc@{}}
    \toprule
    \multicolumn{1}{c}{\textbf{Geo. error ($\times$100)}}      & \multicolumn{1}{c}{\textbf{SURREAL-PV}}
    \\ \midrule

    \multicolumn{1}{l}{DPFM sup~\cite{attaiki2021dpfm}}  & \multicolumn{1}{c}{7.8} \\
    \multicolumn{1}{l}{DPFM unsup~\cite{attaiki2021dpfm}}  & \multicolumn{1}{c}{12.0}  \\
    \multicolumn{1}{l}{Ours}  & \multicolumn{1}{c}{\textbf{6.3}} \\ \hline
    \end{tabular} 
    \vspace{-2mm}
    \caption{Quantitative results on the SURREAL partial view dataset. The \textbf{best} results is highlighted. Our method outperforms both supervised and unsupervised DPFM.}
    \label{tab:partial_view}
\end{table}

\begin{figure}[hbt!]
    \centering
    \includegraphics[width=\linewidth]{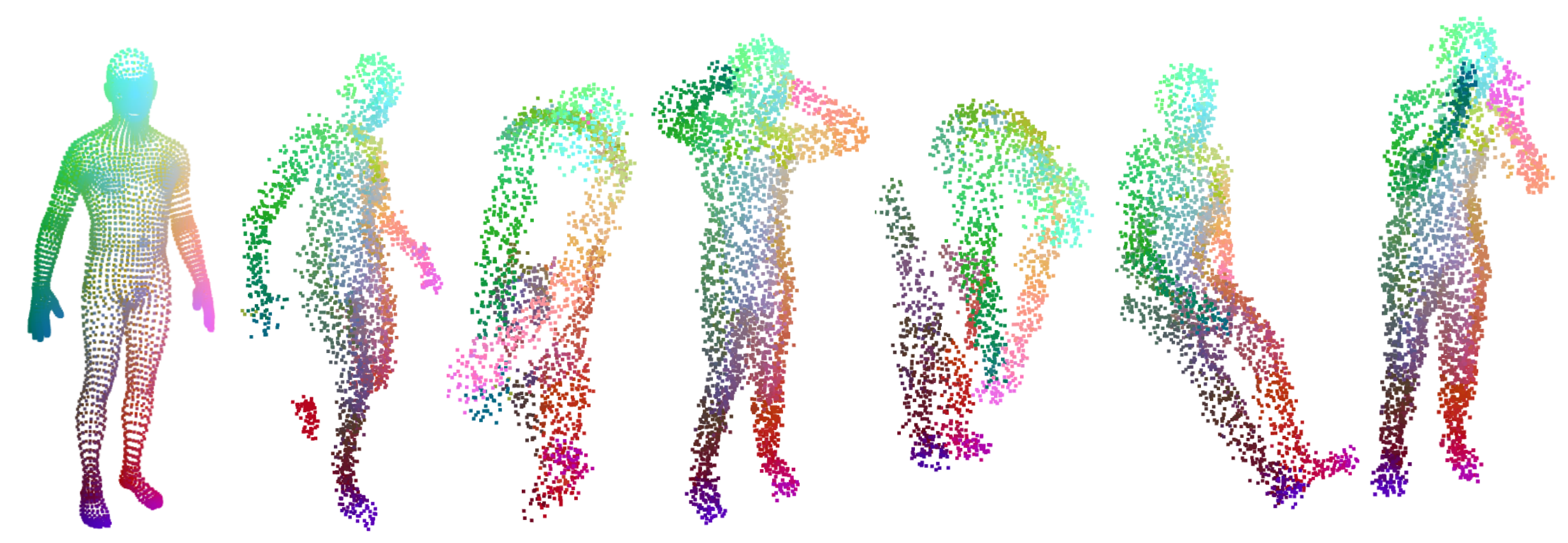}
     \vspace{-7mm}
    \caption{Qualitative results of our method on the SURREAL-PV dataset. The leftmost shape is the source shape. {Our method obtains accurate correspondences even for partially-observed point clouds with different sampling and disconnected components.}}
    \label{fig:partial_view}
\end{figure}

\noindent \textbf{Results.} The results are shown in~\cref{tab:partial_view}. We compare our method with DPFM~\cite{attaiki2021dpfm}, which is the state-of-the-art partial matching method (see~\cref{tab:partial_comparison}). Our method even outperforms the supervised version of DPFM.~\cref{fig:partial_view} shows some qualitative results of our method.

\subsection{Multimodal medical shape data}
\begin{figure}[hbt!]
    \centering
    \includegraphics[width=\linewidth]{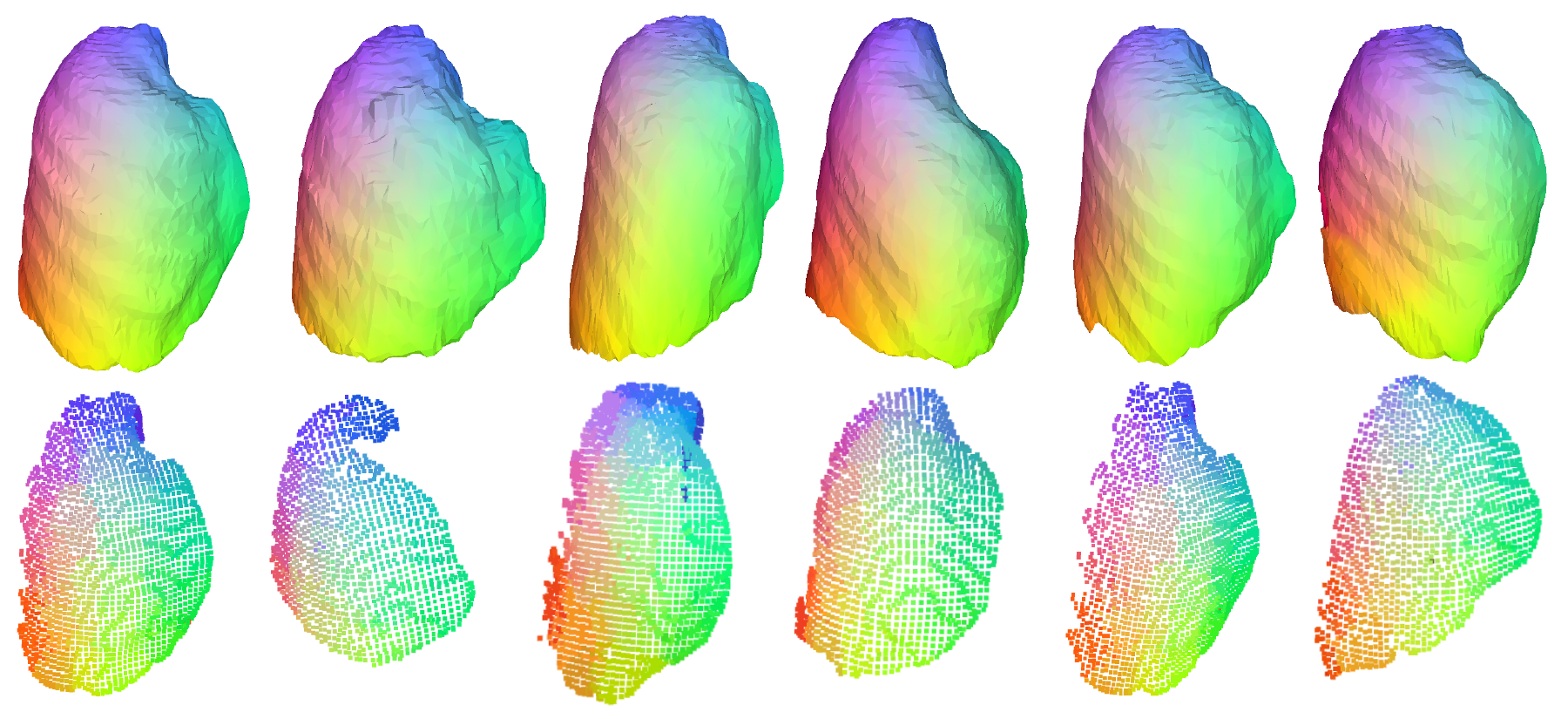}
     \vspace{-7mm}
    \caption{We obtain reliable correspondences for the matching of meshes and partial point clouds of 3D lung shapes.}
    \label{fig:lung}
\end{figure}

To demonstrate the potential for real-world applications, we conduct an experiment for multimodal matching in the context of medical image analysis.
To this end, we use the real-world LUNA16 dataset~\cite{setio2017validation}, which provides chest CT-scans with corresponding lung segmentation mask. Based on the provided segmentation masks, we reconstruct 3D lung shapes represented as triangle meshes and simulate partial point clouds using a subset of  slices of the volumetric images. Since there are no ground-truth correspondences available, we restrict ourselves to qualitative results, which are shown in~\cref{fig:lung}, where it can be seen that reliable correspondences between different lung shapes can be obtained from our method.

\section{Ablation study}
\label{sec:ablation_study}
We evaluate the importance of our introduced loss terms  $E_{\mathrm{align}}$ in~\cref{eq:align_unsup} and $E_{\mathrm{nce}}$ in~\cref{eq:nce} by discarding them individually. For this experiment, we consider the same experimental setting as in~\cref{subsec:complete_shape}. More ablative experiments are provided in the supplementary document.
\begin{table}[hbt!]
    \centering
    \footnotesize
    \begin{tabular}{@{}lccc@{}}
    \toprule
    \multicolumn{1}{c}{\textbf{Geo.~error ($\times$100)}}  & \multicolumn{1}{c}{\textbf{F (PC)}} & \multicolumn{1}{c}{\textbf{S (PC)}} & \multicolumn{1}{c}{\textbf{S19 (PC)}}
    \\ \midrule
    \multicolumn{1}{l}{w.o. $E_{\mathrm{align}}$, $E_{\mathrm{nce}}$}  & 2.3 (5.2) & 4.8 (9.3) & 4.3 (11.2) \\
    \multicolumn{1}{l}{w.o. $E_{\mathrm{nce}}$}  & 2.1 (2.7) & 4.2 (5.6) &  4.1 (5.1) \\
    \multicolumn{1}{l}{w.o. $E_{\mathrm{align}}$}  & 2.0 (22.7) & 3.2 (20.9) & 4.0 (30.8) \\
    \multicolumn{1}{l}{Ours}  & \textbf{2.0} (\textbf{2.4}) & \textbf{3.1} (\textbf{4.1}) & \textbf{4.0} (\textbf{4.5}) \\ \hline
    \end{tabular} 
    \vspace{-2mm}
    \caption{Ablation study on the \textbf{F}AUST, \textbf{S}CAPE and \textbf{S}HREC'\textbf{19} datasets. The \textbf{best} results in each column are highlighted.}
    \label{tab:unsupervised_loss}
\end{table}

\noindent \textbf{Results.} \cref{tab:unsupervised_loss} summarises the quantitative results. By comparing the first row with the second row, we can conclude that $E_{\mathrm{align}}$ plays the key role for accurate point cloud matching. By comparing the first row with the third row, we notice that $E_{\mathrm{nce}}$ can boost the matching performance for triangle meshes, while it hampers the matching performance for point clouds. Together with both loss terms, our method achieves {accurate multimodal matchings}.

\section{Limitations and discussion}
Our work is the first self-supervised approach for multimodal 3D non-rigid shape matching and achieves state-of-the-art performance on a diverse set of relevant tasks. Yet, there are also some limitations that give rise to interesting future research questions.

Unlike previous deep functional map methods that only work for noise-free meshes, our method can handle both meshes and point clouds, even under noise and partiality. However, our method struggles with severe  outliers, {since our method does not have an explicit outlier rejection mechanism and assumes that vertices on the shape with fewer vertices always have a correspondence on the other}.

Analogous to many learning-based shape matching approaches, our method takes 3D vertex positions as input, and is thus not rotation-invariant. However, unlike deformation-based methods~\cite{groueix20183d,zeng2021corrnet3d,sundararaman2022implicit}, {which predict coordinate-dependent deformation fields and thus require rigidly-aligned shapes}, our method allows for data augmentation by randomly rotating shapes during training (similar to~\cite{donati2020deep,attaiki2021dpfm}), so that it is more robust to the initial pose, see the supplementary document for an experimental evaluation.

\section{Conclusion}
In this work we propose the first self-supervised learning framework for multimodal non-rigid shape matching. Our method allows to compute intramodal correspondences for meshes, complete point clouds, and partial point clouds, as well as correspondences across these modalities.
This is achieved by introducing a novel multimodal training strategy that combines mesh-based functional map regularisation with self-supervised contrastive learning coupling mesh and point cloud data. We experimentally demonstrate that our method achieves state-of-the-art performance on numerous benchmarks in diverse settings, including complete shape matching, cross-dataset generalisation, partial shape and partial view matching, as well as multimodal medical shape matching. Overall, we believe that our method will be a valuable contribution towards bridging the gap between the theoretical advances of shape analysis and its practical application in real-world settings, in which partially observed and multimodal data plays an important role.

\subsection*{Acknowledgements}
We thank Zorah Lähner for helpful feedback.
This project was supported by the Deutsche Forschungsgemeinschaft (DFG, German Research Foundation) – 458610525.
This work was supported by the Visual Computing Incubator at the University of Bonn.

{\small
\bibliographystyle{ieee_fullname}
\bibliography{egbib}
}

\section{Supplementary Document}
In this supplementary document we first provide more implementation details of our method. Next, we explain details on the modifications of our approach for partial shape matching. Afterwards, we provide more ablative experiments to demonstrate the advantages of our method. Eventually, we show additional qualitative results of our method.
 
\subsection{More implementation details}
Our learning framework is implemented in PyTorch and uses the original  DiffusionNet implementation\footnote{\url{https://github.com/nmwsharp/diffusion-net}}. In the context of the functional map framework, we choose the first 80 LBO eigenfunctions as  basis functions for complete shape matching. For partial shape matching, we choose the number to be 50 and 30 for the CUTS and HOLES subsets of the SHREC’16, respectively. Similarly, we choose the number to be 30 for the partial view matching. As for deep feature similarity, we use Sinkhorn normalisation with the number of iterations equal to $10$ and temperature parameter equal to $0.2$. We train our feature extractor with the Adam optimiser with learning rate equal to $10^{-3}$. The batch size is chosen to be 8 for SURREAL dataset and 1 for other datasets.

\subsection{Modifications for partial shape matching}
In the context of partial shape matching, the functional map from the complete shape to the partial shape becomes a slanted diagonal matrix. Analogous to DPFM, we regularise the predicted functional maps based on this property. Specifically, for $\mathcal{X}$ being the complete shape and $\mathcal{Y}$ being the partial shape, the unsupervised functional map regularisation can be modified as
\begin{equation}
     \mathcal{L}_{\mathrm{bij}}=\|C_{xy}C_{yx} - \mathbf{I}_{r}\|_{F}^{2},
     \mathcal{L}_{\mathrm{orth}}=\|C_{xy}C_{xy}^{\top} - \mathbf{I}_{r}\|_{F}^{2},
\end{equation}
where $\mathbf{I}_{r}$ is a diagonal matrix in which the first $r$ elements on the diagonal are equal to 1, and $r$ is related to the surface area ratio between two shapes. To obtain the soft correspondence matrix $\hat{\Pi}_{xy}$, we replace  Sinkhorn normalisation by the column-wise softmax operator.

\subsection{Ablation study}

\paragraph{Supervised contrastive learning.} One of the key components of our self-supervised loss terms is the unsupervised functional map regularisation. To evaluate its importance, we replace the unsupervised losses in~\cref{eq:fmaps}  and~\cref{eq:align_unsup} by supervised contrastive loss similar to~\cref{eq:nce}, i.e.
    \begin{equation}
        \label{eq:sup_nce}
        E_{\mathrm{sup}} = -\sum_{(i,j)\in\mathcal{P}} \log \frac{\exp \left( \mathcal{F}_{x}^{i} \cdot \mathcal{F}_{y}^{j} / \tau \right)}{\sum_{(\cdot,k)\in\mathcal{P}} \exp \left( \mathcal{F}_{x}^{i} \cdot \mathcal{F}_{y}^{k} / \tau \right)},
    \end{equation}
where $\mathcal{P}$ is the set of matched points between shape $\mathcal{X}$ and shape $\mathcal{Y}$. For this ablation experiment, we consider the same experiment setting as in~\cref{subsec:cross-dataset} to avoid over-fitting.

\begin{table}[hbt!]
    \small
    \centering
        \begin{tabular}{@{}lccc@{}}
        \toprule
        \multicolumn{1}{c}{\textbf{Geo. ($\times$100)}}  & \multicolumn{1}{c}{\textbf{F (PC)}} & \multicolumn{1}{c}{\textbf{S (PC)}} & \multicolumn{1}{c}{\textbf{S19 (PC)}}
        \\ \midrule
        \multicolumn{1}{l}{with $E_{\mathrm{sup}}$}  & \multicolumn{1}{c}{\textbf{1.5} (4.8)} & \multicolumn{1}{c}{5.2 (6.8)} & \multicolumn{1}{c}{6.9 (8.1)} \\
        \multicolumn{1}{l}{Ours} & \multicolumn{1}{c}{2.0 (\textbf{3.5})}  & \multicolumn{1}{c}{\textbf{3.2} (\textbf{3.8})} & \multicolumn{1}{c}{\textbf{4.4} (\textbf{6.6})} \\ \hline
        \end{tabular} 
    \caption{Quantitative results on the \textbf{F}AUST, \textbf{S}CAPE and \textbf{S}HREC'\textbf{19} datasets trained on SURREAL dataset. The \textbf{best} results in each column are highlighted.}
    \label{tab:sup_nce}
\end{table}
The quantitative results are summarised in~\cref{tab:sup_nce}. Notably, our self-supervised approach outperforms the supervised counterpart in most settings. The reason is that the unsupervised functional map regularisation enforces more smooth and consistent correspondences in comparison to the supervised contrastive learning. 

\paragraph{Robustness to initial pose.}
As indicated in the limitation part, our method takes vertex position as input and is thus not rotation-invariant. To be more robust to the choice of initial pose, during training we randomly rotate input shapes as data augmentation, thereby encouraging that the extracted features are less sensitive to the initial pose of the shape. To evaluate the performance, we follow the experiment setting in~\cref{subsec:complete_shape}, with the only  difference being that here all test shapes are randomly rotated around the vertical axis. 

\begin{table}[hbt!]
    \small
    \centering
    \begin{tabular}{@{}lccc@{}}
        \toprule
        \multicolumn{1}{c}{\textbf{Geo. ($\times$100)}}  & \multicolumn{1}{c}{\textbf{F (PC)}} & \multicolumn{1}{c}{\textbf{S (PC)}} & \multicolumn{1}{c}{\textbf{S19 (PC)}}
        \\ \midrule
        \multicolumn{1}{l}{Ours (w/o aug.)}  & \multicolumn{1}{c}{{8.8} (12.0)} & \multicolumn{1}{c}{14.0 (14.2)} & \multicolumn{1}{c}{13.9 (14.7)} \\
        \multicolumn{1}{l}{Ours (w/ aug.)} & \multicolumn{1}{c}{\textbf{4.7} (\textbf{5.6})}  & \multicolumn{1}{c}{\textbf{5.3} (\textbf{6.2})} & \multicolumn{1}{c}{\textbf{6.0} (\textbf{6.8})} \\ \hline
    \end{tabular} 
    \caption{Quantitative results on the \textbf{F}AUST, \textbf{S}CAPE and \textbf{S}HREC'\textbf{19} datasets in terms of mean geodesic errors (×100). All test shapes are randomly rotated. The best results in each column are highlighted.}
    \label{tab:initial_pose}
\end{table}

\begin{figure}[hbt!]
    \centering
    \includegraphics[width=\linewidth]{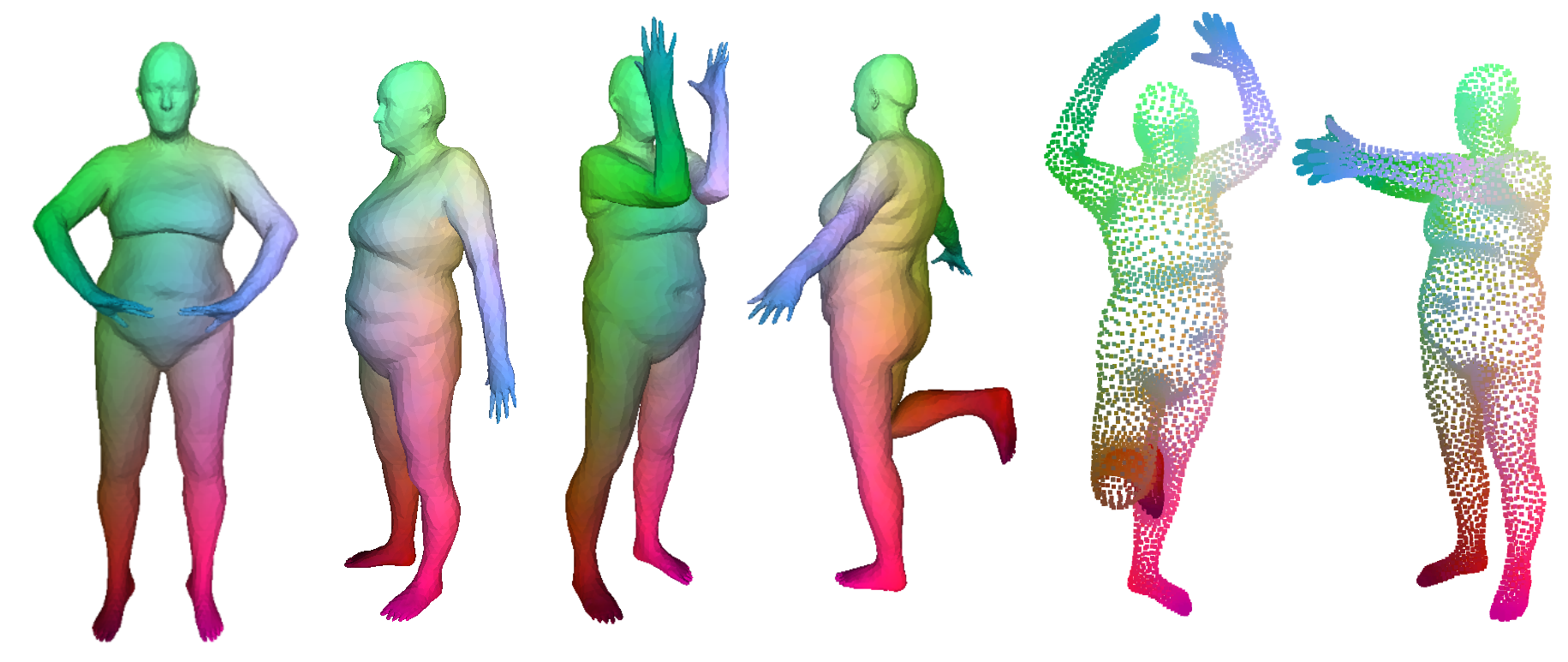}
    \caption{Qualitative results on FAUST dataset with different initial poses for both mesh and point cloud matching.}
    \vspace{-8mm}
    \label{fig:faust_rot}
\end{figure}

\cref{tab:initial_pose} summarises the quantitative results. We observe that the network performance can be substantially improved by using a random rotation as data augmentation during training. \cref{fig:faust_rot} shows some qualitative results of our method on FAUST dataset with randomly initial poses.

\paragraph{Robustness to noise.} As mentioned in the main paper, previous deep functional map methods predict point maps based on the functional map framework. However, point clouds only admit an inaccurate estimation of LBO eigenfunctions, especially in the presence of noise. Therefore, directly applying such methods to point clouds leads to a large performance drop. In contrast, our method predicts point maps based on the deep feature similarity without relying on the functional map framework.

\begin{figure}[hbt!]
    \centering
    \includegraphics[width=\linewidth]{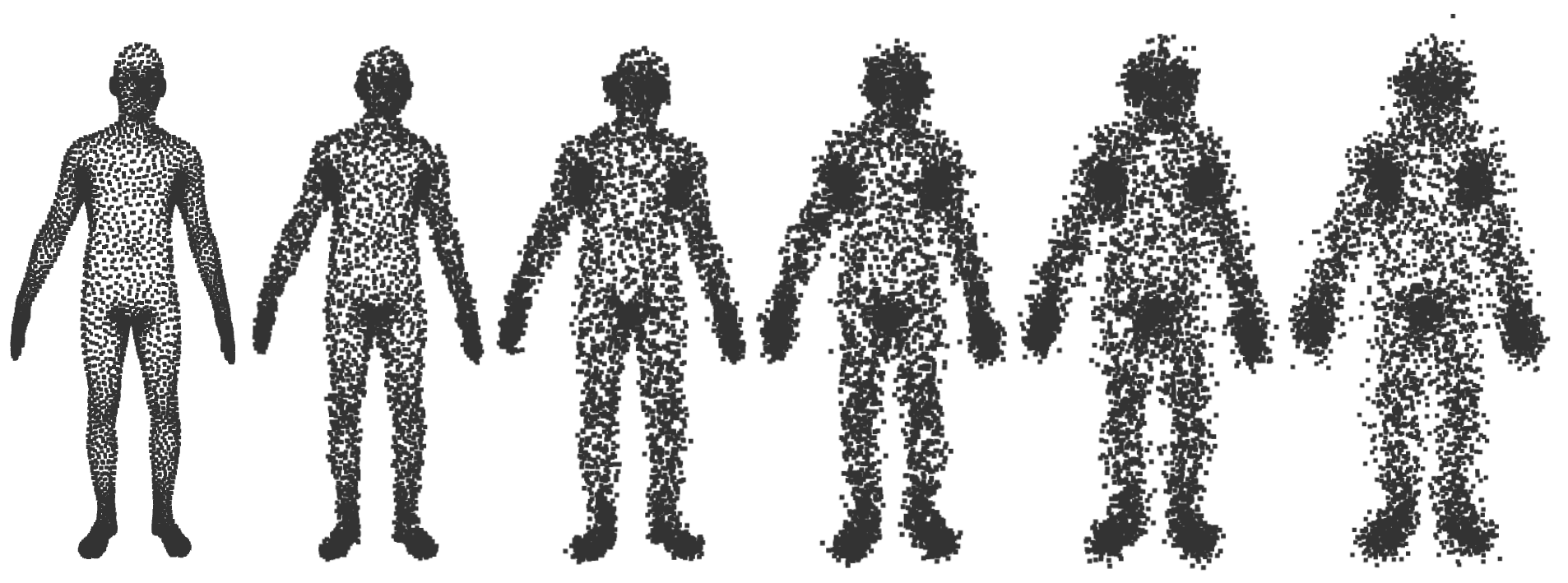}
    \caption{One shape on FAUST dataset with increasing noise magnitude from left to right. The leftmost one is the clean point cloud.}
    \vspace{-3mm}
    \label{fig:noise_scale}
\end{figure}

\begin{figure}[hbt!]
    \centering
    \includegraphics[width=\linewidth]{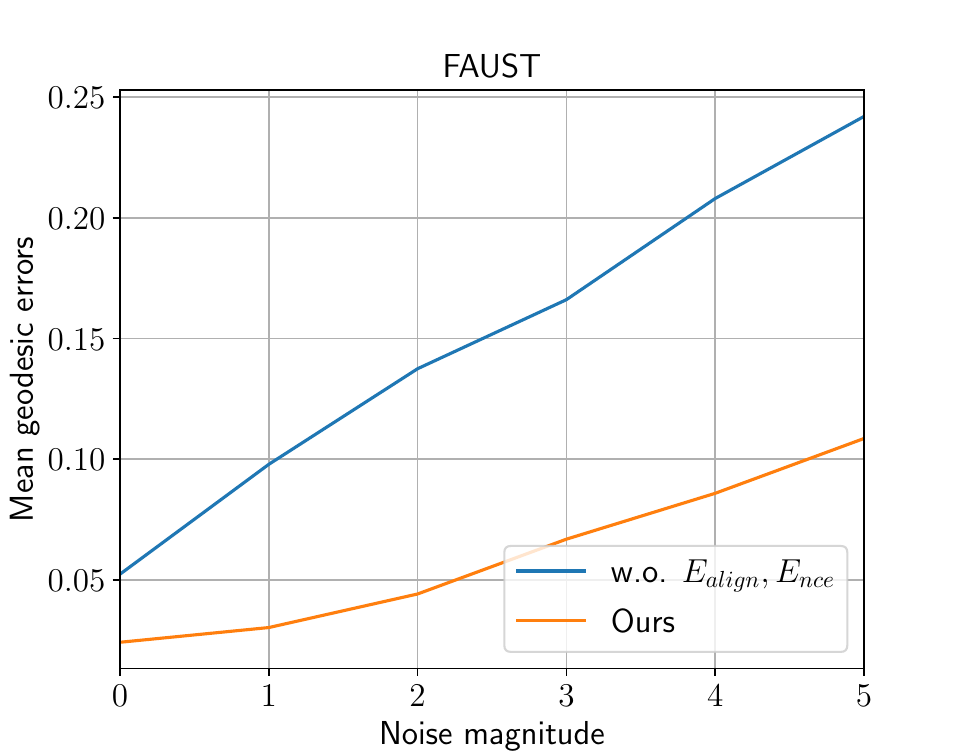}
    \caption{Mean geodesic errors for point cloud matching in different noise magnitudes. Our method achieves more robust point cloud matching based on deep feature similarity.}
    \label{fig:noise}
    \vspace{-3mm}
\end{figure}

To further evaluate our method's robustness to noise, we add an increasing amount of zero-mean isotropic Gaussian noise to point positions of shapes in the test set of the FAUST dataset. For a fair comparison, we do not train or fine-tune the networks on each noise magnitude. As a proof-of-concept, we choose our method and a simple baseline that is based on our framework but does not use  $E_{align}, E_{nce}$ during training for comparison, which is similar to~\cref{sec:ablation_study}. We note that the simple baseline predicts point maps based on the functional map framework via functional maps conversion.~\cref{fig:noise_scale} shows the point cloud with different noise magnitude.~\cref{fig:noise} plots the mean geodesic errors on the FAUST dataset w.r.t. the corresponding noise magnitude. We observe that our method achieves better results and is much more robust against noise, especially for large degrees of noise.

\paragraph{Robustness to sampling.} To evaluate our method's robustness to varying sampling density, we use an anisotropic remeshed version of the FAUST and SCAPE datasets  (denoted F\_a and S\_a). Below we show an example shape pair with varying sampling density and corresponding matching.

\begin{figure}[hbt!]
    \centering
    \begin{tabular}{cccc}
       \includegraphics[height=2.8cm]{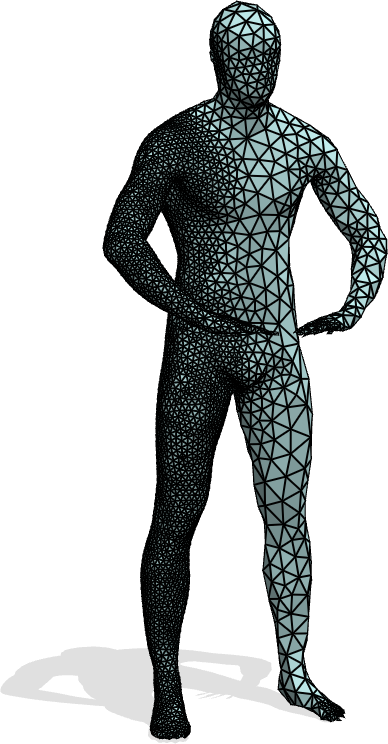}  & \includegraphics[height=2.8cm]{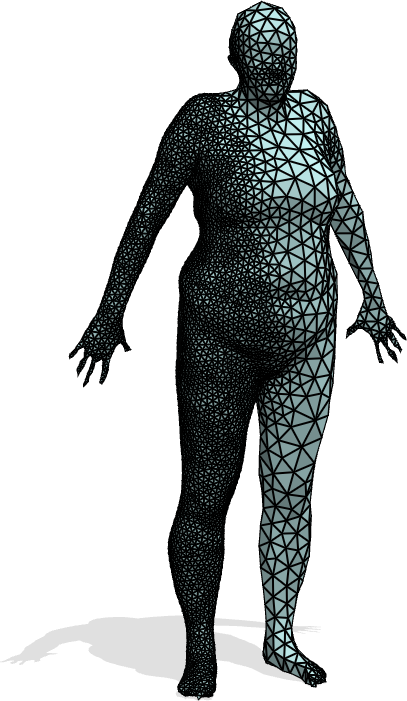}  &
       \includegraphics[height=2.8cm]{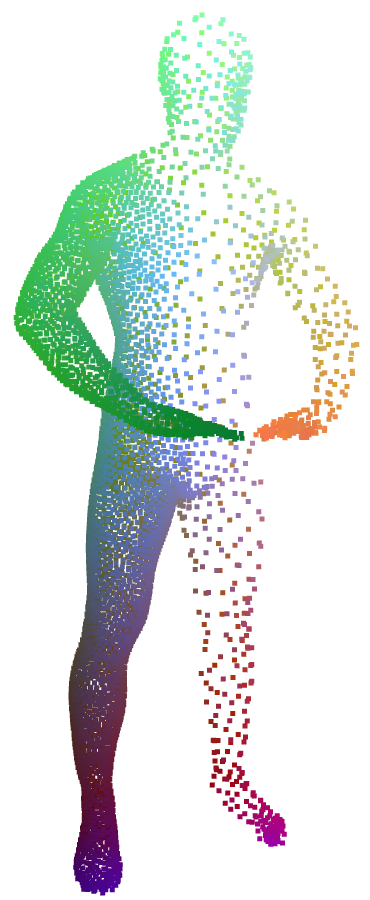} &
      \includegraphics[height=2.8cm]{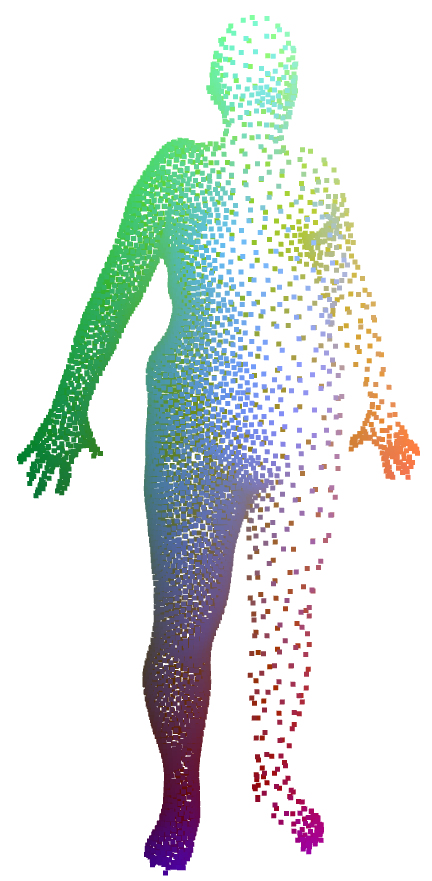}
       \\
    \end{tabular}
    \vspace{-2mm}
    \caption{An example shape pair and the corresponding qualitative result of our method on the anisotropic remeshed FAUST dataset.}
    \label{fig:aniso}
\end{figure}

\begin{table}[hbt!]
\setlength{\tabcolsep}{3.5pt}
    \footnotesize
    \centering
    \begin{tabular}{@{}lcccc@{}}
    \toprule
    \multicolumn{1}{l}{Train}  & \multicolumn{2}{c}{\textbf{FAUST}}   & \multicolumn{2}{c}{\textbf{SCAPE}}\\ \cmidrule(lr){2-3} \cmidrule(lr){4-5}
    \multicolumn{1}{l}{Test} & \multicolumn{1}{c}{\textbf{F (PC)}} & \multicolumn{1}{c}{\textbf{F\_a (PC)}} & \multicolumn{1}{c}{\textbf{S (PC)}} & \multicolumn{1}{c}{\textbf{S\_a (PC)}}
    \\ \midrule
    \multicolumn{1}{l}{ConsistFMaps (w/ xyz)} & \multicolumn{1}{l}{2.4 (11.2)} & \multicolumn{1}{l}{2.9 (12.4)} & \multicolumn{1}{l}{5.1 (12.3)} & \multicolumn{1}{l}{5.4 (13.1)}\\
    \multicolumn{1}{l}{~~~~~~w/ SHOT (original)} & \multicolumn{1}{l}{\textbf{1.5} (16.4)} & \multicolumn{1}{l}{15.3 (32.1)} & \multicolumn{1}{l}{\textbf{2.0} (18.3)} & \multicolumn{1}{l}{6.9 (24.8)}\\
    \multicolumn{1}{l}{Deep Shells (w/ xyz)} & \multicolumn{1}{l}{1.7 (6.0)} & \multicolumn{1}{l}{2.7 (7.2)} & \multicolumn{1}{l}{5.3 (7.8)} & \multicolumn{1}{l}{5.7 (8.4)}\\
    \multicolumn{1}{l}{~~~~~~w/ SHOT (original)} & \multicolumn{1}{l}{1.7 (13.2)} & \multicolumn{1}{l}{12.0 (18.8)} & \multicolumn{1}{l}{2.5 (14.1)} & \multicolumn{1}{l}{10.0 (18.3)}\\
    \multicolumn{1}{l}{Ours} & \multicolumn{1}{l}{2.0 (\textbf{2.4})} & \multicolumn{1}{l}{\textbf{2.6} (\textbf{3.0})} & \multicolumn{1}{l}{3.1 (\textbf{4.1})} & \multicolumn{1}{l}{\textbf{3.3} (\textbf{4.4})}\\
    \hline
    \end{tabular}
    \caption{Quantitative results on \textbf{F}AUST, \textbf{S}CAPE and their anisotropic remeshed versions. All methods are trained on the original datasets. }
    \label{tab:anisotropic}
\end{table}
Tab.~\ref{tab:anisotropic} shows that 
both ConsistFMaps and Deep Shells overfit to the sampling density
(with SHOT descriptor they overfit both for meshes and point clouds; with vertex position as descriptor they only overfit for point clouds). 
In contrast, our method is more robust  and demonstrates better performance, particularly for point clouds. 

\begin{figure}[hbt!]
    \centering
    \begin{tabular}{cccc}\includegraphics[height=2.8cm]{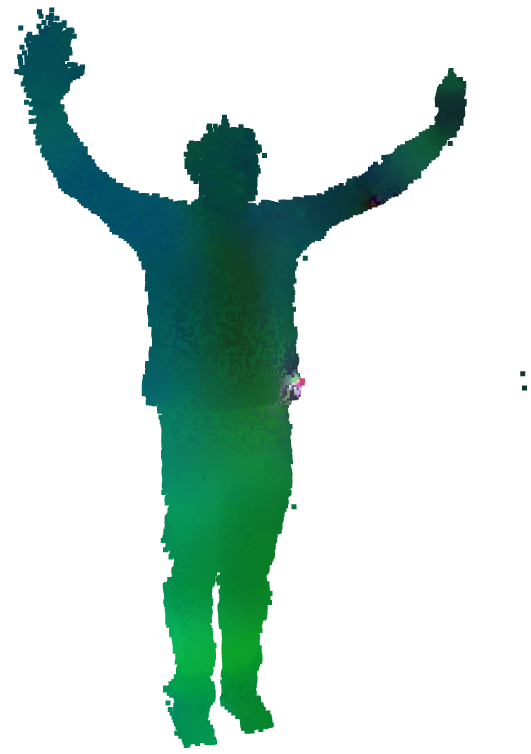}  &
       \includegraphics[height=2.8cm]{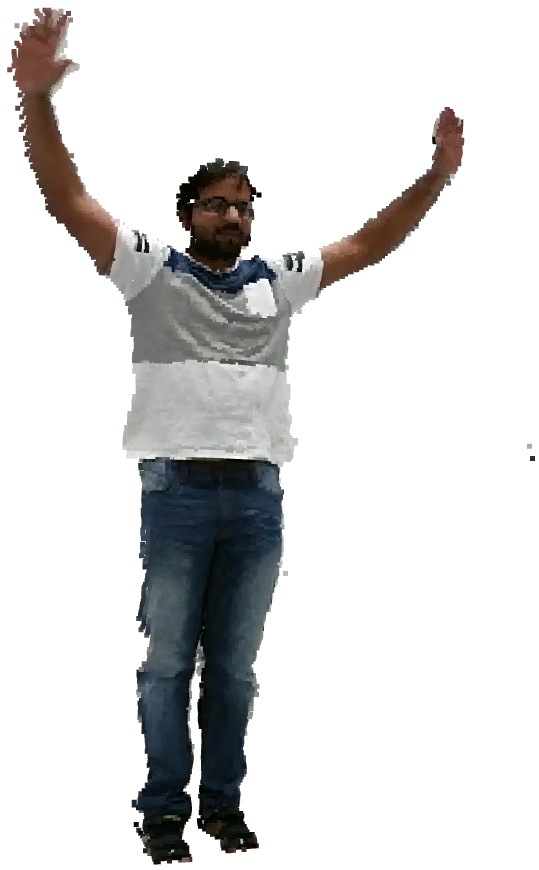}  & 
       \includegraphics[height=2.8cm]{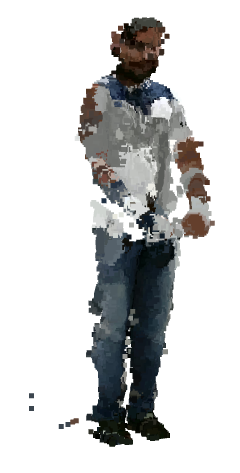} &
      \begin{overpic}[height=2.8cm]{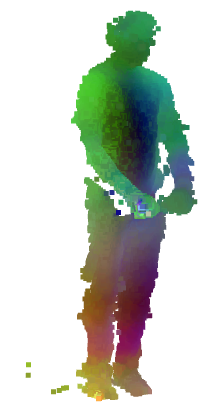}
      \put(20,40){\includegraphics[height=0.5cm]{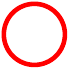}}
      \end{overpic}
       \\
    \end{tabular}
    \caption{A qualitative result from real-scanned raw point clouds and the corresponding extracted features from DiffusionNet.}
    \vspace{-3mm}
    \label{fig:outlier}
\end{figure}

 \paragraph{Matching with outliers.}~\cref{fig:outlier} shows an example matching result from real-scanned raw point clouds (by transferring texture). We observe  that the extracted DiffusionNet features (see colour-coded shapes on the left and right, which visualise DiffusionNet features projected onto three RGB channels via t-SNE) are degraded due to the outliers. Since we use DiffusionNet, our method carries over this known limitation (of DiffusionNet).

\paragraph{Data efficiency.} We train our method on the entire SURREAL dataset and summarise the results in~\cref{tab:data}. When using significantly more data, our method achieves a (slightly) better cross-dataset generalisation ability. Compared to point cloud matching methods, our method utilises the strong functional map regularisation and explicitly considers multi-modal training, thus requires only a small amount of training data.
\begin{table}[hbt!]
    \footnotesize
    \centering
    \begin{tabular}{@{}lccc@{}}
    \toprule
    \multicolumn{1}{c}{\textbf{$|$Data$|$}} & \multicolumn{1}{c}{\textbf{F (PC)}}   & \multicolumn{1}{c}{\textbf{S (PC)}}  & \multicolumn{1}{c}{\textbf{S19 (PC)}}  \\\midrule
    \multicolumn{1}{c}{5k}  & \multicolumn{1}{c}{{2.0} ({3.5})}  & \multicolumn{1}{c}{{3.2} ({3.8})} & \multicolumn{1}{c}{{4.4} ({6.6})} \\
    \multicolumn{1}{c}{230k}  & \multicolumn{1}{c}{\textbf{1.9} (\textbf{3.2})}  & \multicolumn{1}{c}{\textbf{3.0} (\textbf{3.6})} & \multicolumn{1}{c}{\textbf{4.0} (\textbf{5.8})} \\
    \hline
    \end{tabular}
    \caption{Cross-dataset generalisation evaluated on the \textbf{F}AUST, \textbf{S}CAPE and \textbf{S}HREC’\textbf{19} datasets and trained on the SURREAL dataset.}   
    \label{tab:data}
\end{table} 

\subsection{More qualitative results}
\begin{figure}[hbt!]
    \centering
    \includegraphics[width=\linewidth]{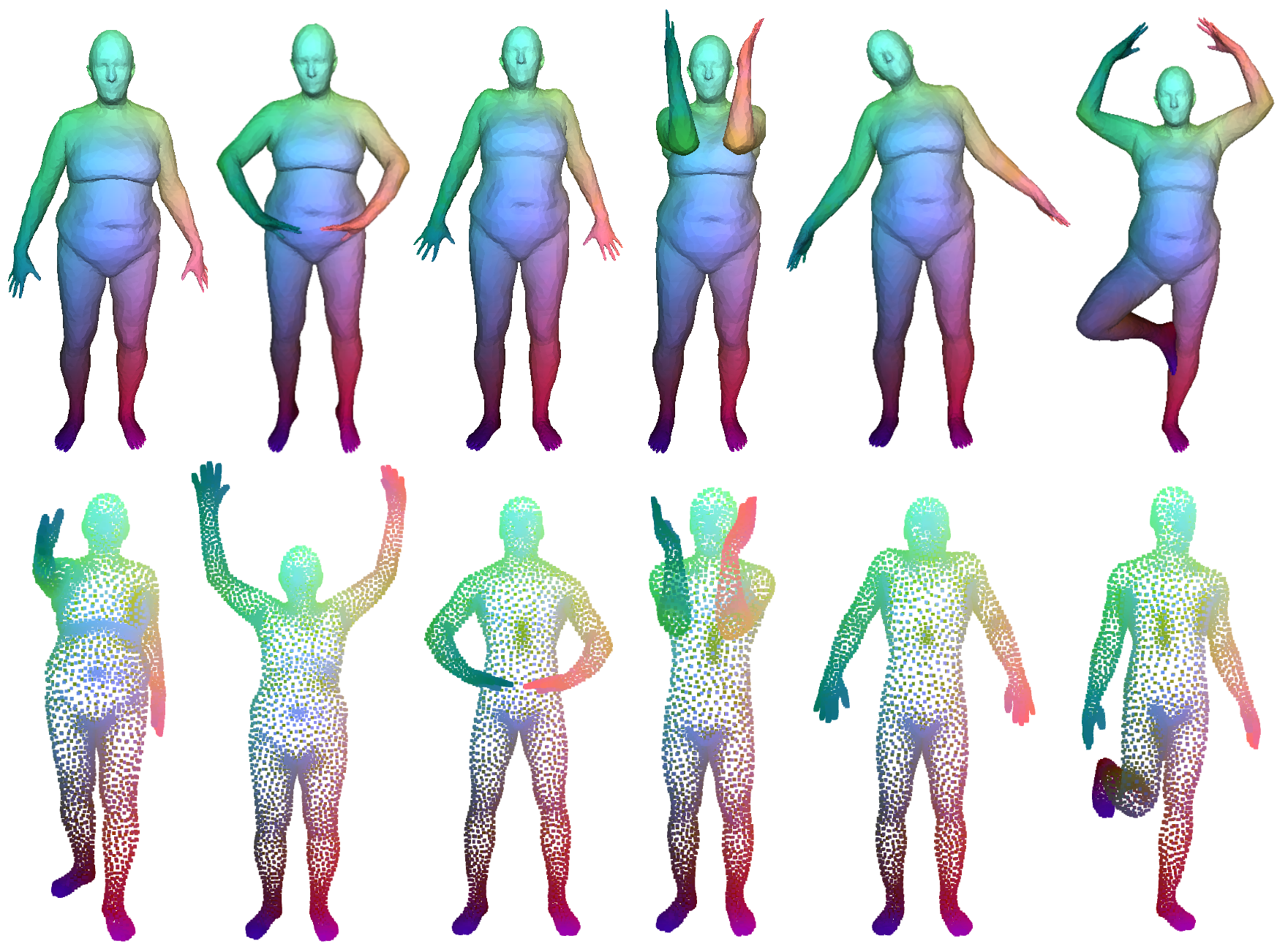}
    \caption{Qualitative results on FAUST dataset of our method applied to both meshes and point clouds. Our method achieves accurate matchings for both modalities.}
    \label{fig:faust_vis}
\end{figure}

\begin{figure}[hbt!]
    \centering
    \includegraphics[width=0.95\linewidth]{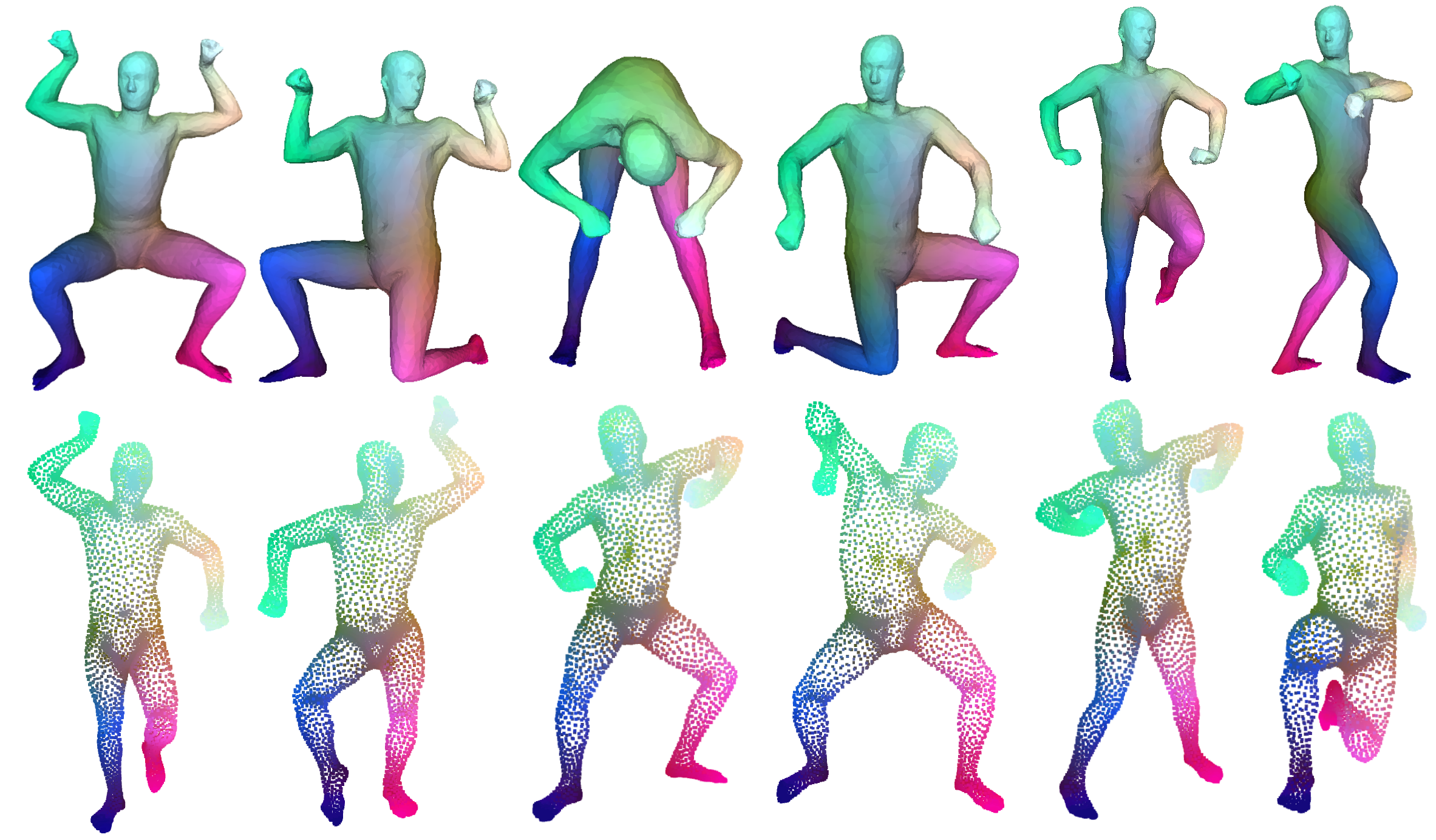}
    \caption{Qualitative results on SCAPE dataset of our method applied to both meshes and point clouds.}
    \label{fig:scape_vis}
\end{figure}

\begin{figure}[hbt!]
    \centering
    \includegraphics[width=0.95\linewidth]{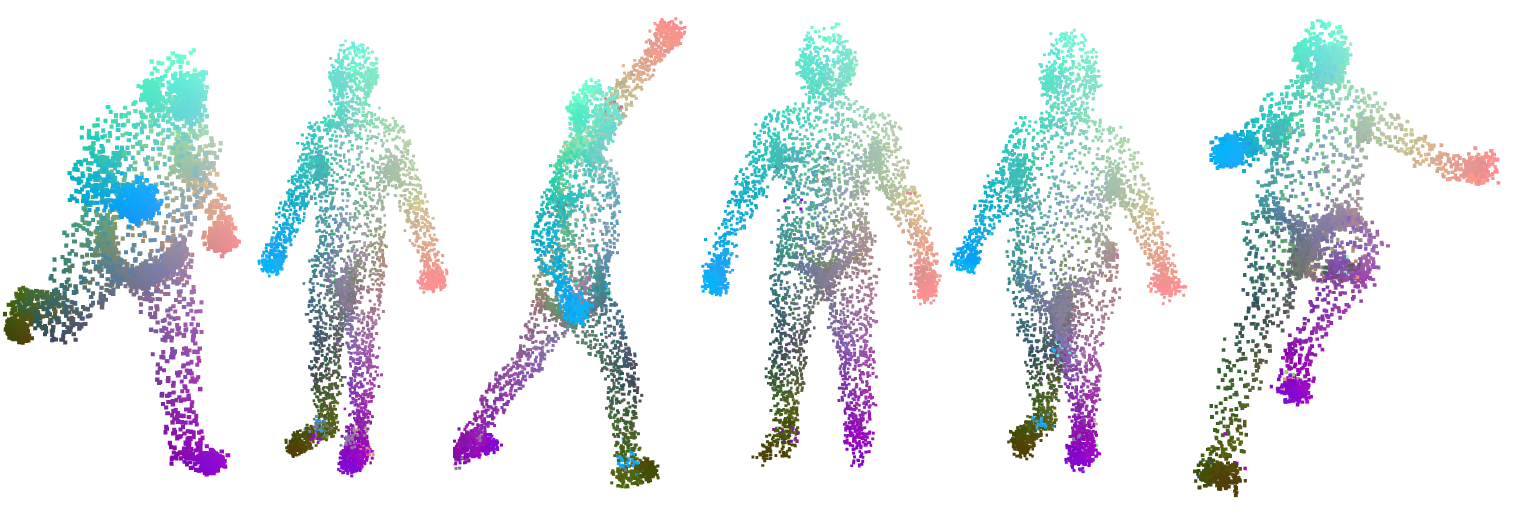}
    \caption{Qualitative results on SHREC'19 dataset of our method applied to noisy point clouds. Our method enables accurate point cloud matching even in the presence of noise.}
    \label{fig:shrec19_noise}
\end{figure}

\begin{figure}[hbt!]
    \centering
    \includegraphics[width=\linewidth]{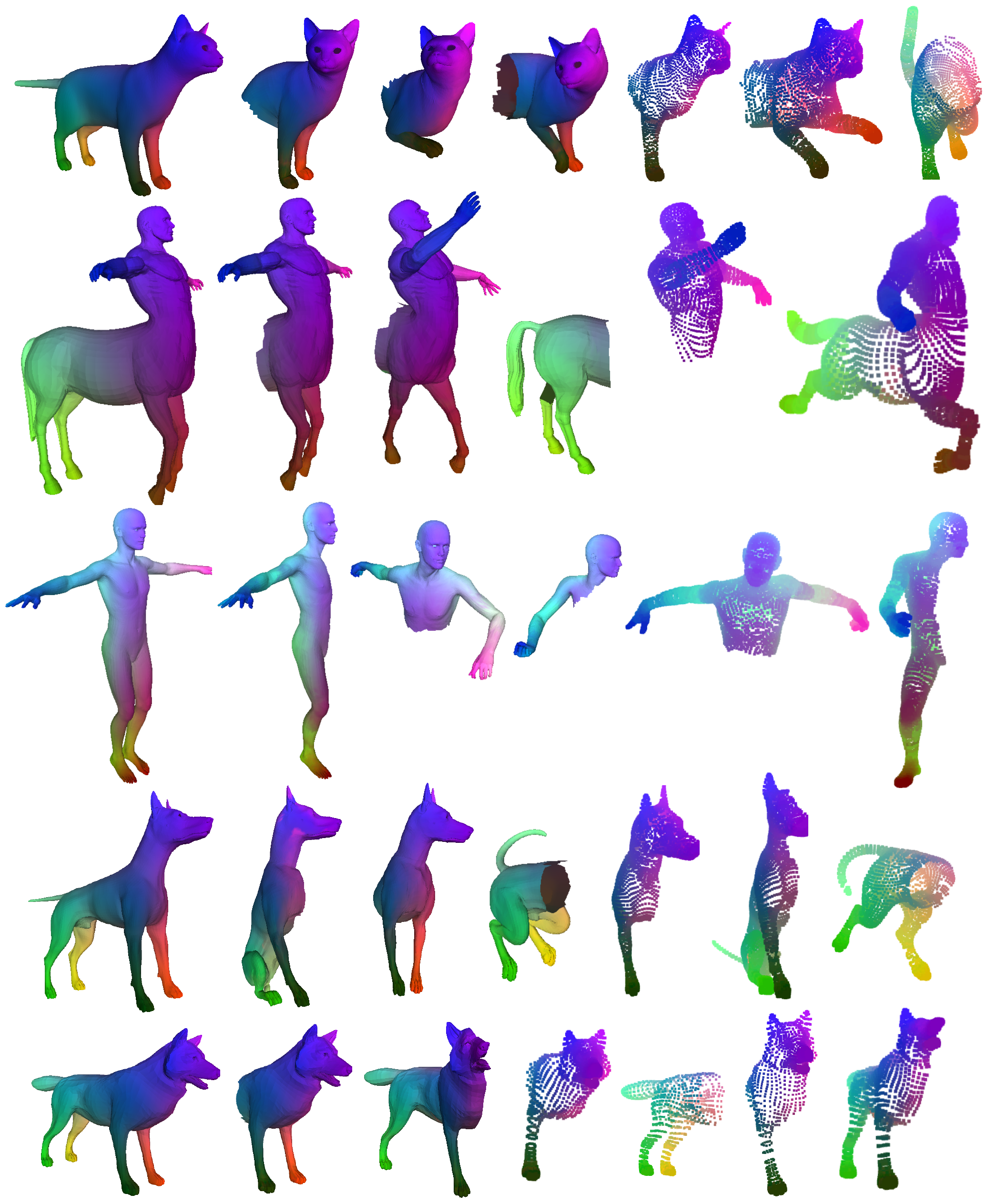}
    \caption{Qualitative partial shape matching results on SHREC'16 dataset of our method applied to both meshes and point clouds. The leftmost one is the complete shape to be matched. Our method enables accurate multimodal partial shape matching.}
    \label{fig:shrec16_vis}
\end{figure}

\begin{figure}[hbt!]
    \centering
    \includegraphics[width=0.96\linewidth]{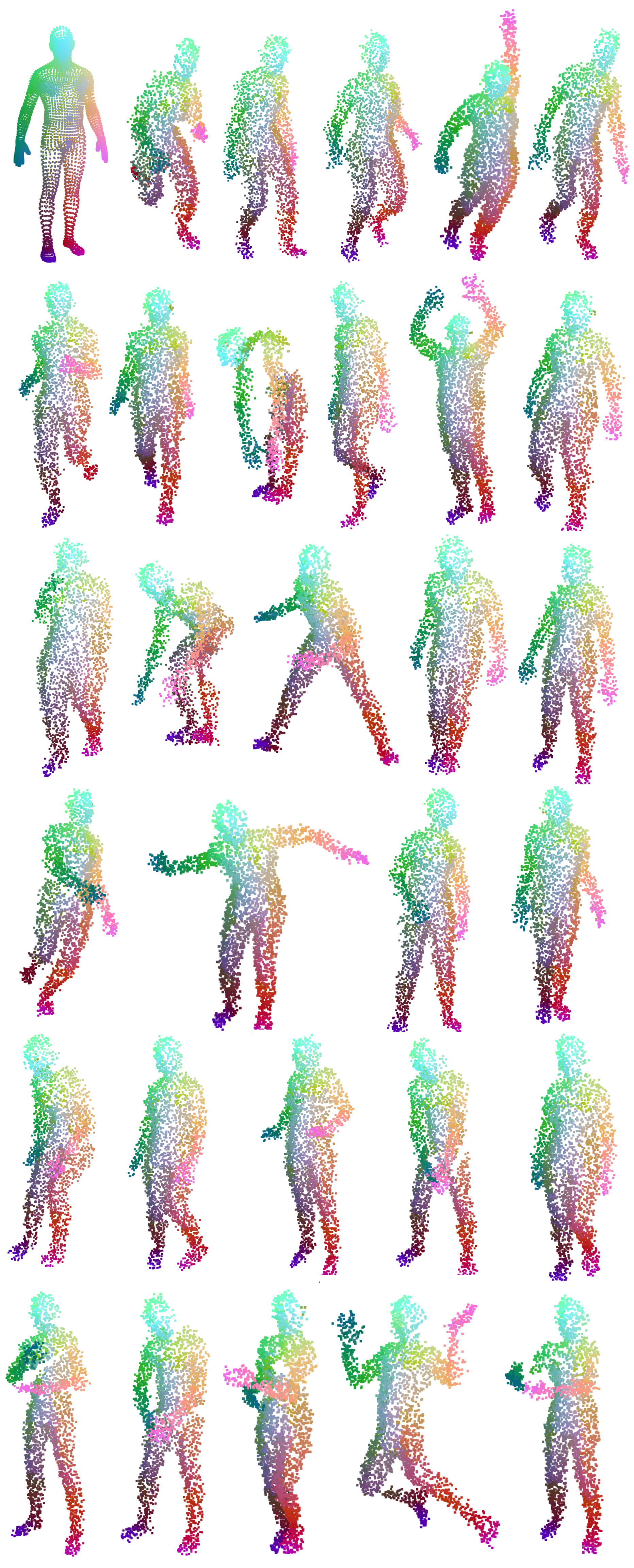}
    \caption{Qualitative partial view matching results on SURREAL-PV dataset of our method applied to noisy partial observed point clouds. The top-left one is the complete shape to be matched. Our method obtains accurate correspondences for partially-observed noisy point clouds with different sampling and disconnected components.}
\label{fig:surreal_partial_view}
\end{figure}

In this section, we provide more qualitative matching results on diverse shape matching datasets, see Figs.~\ref{fig:faust_vis}-\ref{fig:surreal_partial_view}.

\end{document}